\documentclass[12pt,twoside]{article}
\usepackage[margin=2.38cm]{geometry}
\usepackage{graphicx}
\usepackage{amsmath}
\usepackage[T1]{fontenc}
\usepackage[utf8]{inputenc}

\usepackage[misc]{ifsym}

\usepackage[hyphens]{url}

\usepackage[style=apa,backend=biber]{biblatex}
\usepackage{hyperref}

\usepackage{float}

\usepackage{xcolor}

\usepackage[most]{tcolorbox}
\tcbset{
  abstractbox/.style={
    colback=gray!5,
    colframe=gray!5,
    boxrule=0.3pt,
    arc=1mm,
    left=6pt, right=6pt, top=4pt, bottom=4pt,
    boxsep=5pt
  },
  expbox/.style={
    enhanced,
    colback=gray!5,
    colframe=gray!30,
    boxrule=0.4pt,
    arc=1mm,
    left=8pt, right=8pt, top=6pt, bottom=6pt,
    boxsep=4pt
  }
}

\definecolor{refcolor}{rgb}{0.1, 0.1, 0.4}

\hypersetup{
    colorlinks=true,
    linkcolor=refcolor,
    citecolor=refcolor,
    urlcolor=refcolor
}

\usepackage[activate={true,nocompatibility},final,auto=true,tracking=true,kerning=true,expansion=true,spacing=true,factor=1000,stretch=20,shrink=20,letterspace=0]{microtype}

\usepackage{crimson}

\usepackage[font=small,labelfont=bf]{caption}

\title{\textbf{Where is the Mind? Persona Vectors and LLM Individuation}}

\usepackage{authblk}

\author[1,2]{Pierre Beckmann}
\author[3]{Patrick Butlin}
\affil[1]{MATS}
\affil[2]{EPFL \& Idiap Research Institute}
\affil[3]{Eleos AI Research}
\affil[ ]{\textcolor{gray}{\small\Letter} \hspace{0.05cm} \href{mailto:pierrebeckmann@gmail.com}{\texttt{pierrebeckmann@gmail.com}}}

\makeatletter
\renewcommand{\maketitle}{\bgroup\setlength{\parindent}{0pt}
\thispagestyle{plain}
\begin{flushleft}
\vspace*{0.65cm}
  {\fontsize{17}{17}\selectfont \@title}
  \vspace{0.4cm}

  {\@author}
\end{flushleft}\egroup
}
\makeatother

\usepackage{titlesec}
\titleformat{\section}
  {\fontsize{17}{15}\selectfont\bfseries}
  {\thesection}{1em}{}

\usepackage{fancyhdr}
\usepackage{lastpage}

\newcommand{\doctitle}{Where is the Mind? Persona Vectors and LLM Individuation}
\newcommand{\docauthors}{Pierre Beckmann and Patrick Butlin}

\fancypagestyle{plain}{%
  \fancyhf{}
  
  \fancyhead[L]{September 2026}
  \fancyhead[R]{Forthcoming in \textit{Ergo}}
}

\graphicspath{{figures/}}

\begin{document}

\maketitle

\vspace{0.15cm}

\begin{tcolorbox}[abstractbox]
\noindent
\textbf{Abstract:}
The individuation problem for large language models asks which entities associated with them, if any, should be identified as minds. We approach this problem through mechanistic interpretability, engaging in particular with recent empirical work on persona vectors, persona space, and emergent misalignment. We argue that three views are the strongest candidates: the virtual instance view and two new views we introduce, the (virtual) instance-persona view and the model-persona view. First, we argue for the virtual instance view on the grounds that attention streams sustain quasi-psychological connections across token-time. Then we present the persona literature, organised around three hypotheses about the internal structure underlying personas in LLMs, and show that the two persona-based views are promising alternatives.

\vspace{0.3cm}
\end{tcolorbox}

\vspace{0.0cm}

\section*{Introduction}
\addcontentsline{toc}{section}{Introduction}

While the initial consensus urged caution against anthropomorphizing LLMs \parencite{bender2020climbing,shanahan2023roleplay}, it has become increasingly difficult to deny them any form of mentality. For one thing, attributing mental states to LLMs, such as beliefs or desires, is remarkably useful for explaining and predicting their behaviour. This fact alone has been taken as a starting point for arguing that LLMs genuinely possess mental states \parencite{goldstein2024mind,goldstein2025chatgpt,cappelen2025wholehog}. Other approaches have focused instead on internal operations, seeking to show that particular representations play the functional roles of particular mental states. On these grounds, it has been argued that LLMs are capable of forming something like beliefs \parencite{herrmann2025standards, slocum2025believe}, credences \parencite{keeling2025confidence}, desires \parencite{butlin2024desire} and intentions \parencite{williams2025intentions}, as well as mental capacities such as understanding \parencite{beckmann2026mechanistic}.

If LLMs form mental-like states, it is natural to ask what entity, if any, these states belong to. This is the individuation question: where should we locate the relevant subject? Is it the model in general, specified by its architecture and weights? Is it a \textit{physical instance}---a GPU running the model at a given time? Or is it a \textit{virtual instance}---the model as it participates in a single conversation, regardless of which hardware realizes each computational step? Or something else entirely? This question is important for understanding mind-like phenomena in LLMs, and in particular for AI welfare. To work out whether LLMs deserve welfare protections, we need to identify the entities that may be moral patients \parencite{long2024taking,register2025individuating}.

A growing philosophical literature has already begun this work \parencite{chalmers2025llm,goldstein2025chatgpt,mcintyre2025individuating,shiller2025digital}, proposing candidates such as the model, the physical instance, and the virtual instance, and articulating criteria for choosing between them. This literature tends to favour the virtual instance and related views. Our first contribution in this paper is to clarify this view and reinforce the arguments in its favour, with reference to the mechanisms of processing in LLMs.

However, the existing literature does not engage sufficiently with a central feature of LLM operation: that they proceed by adopting a context-appropriate persona. An early account of this phenomenon, the \textit{simulators framework}, presented models as being capable of role-playing as any of a wide repertoire of characters \parencite{janus2022simulators,shanahan2023roleplay}. This perspective could be seen as denying that there are individuals associated with LLMs; instead there is the model, which is deeply unlike an individual agent, and the characters, which are ephemeral and mere products of role-play. More recently, there has been a growing recognition that this picture underplays the importance of personas for LLMs' safety and efficacy. A newer perspective sees post-training and system prompts as shaping the persona of a helpful AI assistant; by default, models work by predicting this persona's behaviour \parencite{nostalgebraist2025void,marks2026psm}. Anthropic's approach to post-training makes this particularly clear: it is fundamentally guided by Claude's Constitution, which explicitly describes a character and set of values \parencite{bai2022constitutional,anthropic2026constitution}.

Recent empirical work also supports this new perspective. In cases of \textit{emergent misalignment}, fine-tuning on a narrow task like insecure code systematically pushes models into a broadly coherent `evil' persona that generalizes across unrelated contexts. Work on \textit{persona vectors} explains why: there are directions in activation space that organize dispositional profiles along which the model can easily select the most context-appropriate persona. We can now find the mechanistic bases of coherent, repeatable personas. So our second contribution is to present this research and argue for a role for personas in LLM individuation.

Overall, our aim in this paper is to identify the strongest candidates for LLM individuation. In section~\ref{sec:problem}, we give a more detailed account of the individuation problem and the candidates that have so far been proposed. In section~\ref{sec:mechanistic}, we argue on mechanistic grounds that the virtual instance is the strongest of these; support comes from the fact that attention streams sustain quasi-psychological connections across token-time. In section~\ref{sec:personaspace}, we turn to personas, presenting empirical evidence for three hypotheses about their implementation in LLMs. Finally, in section~\ref{sec:personaviews}, we use these findings to motivate two new candidates---the instance-persona view and the model-persona view---and argue that neither can be easily dismissed. Along the way, we identify open empirical questions and flag them as research directions for mechanistic interpretability.

\section{The problem and proposed solutions}\label{sec:problem}

\subsection{The individuation problem}\label{sec:individuation}

The individuation problem is the problem of saying which processes in LLMs, and outputs thereof, should be attributed to the same minds. Expressed in this way, it arises only if one assumes or suspects that there are minds in LLMs, but this need not imply much metaphysical commitment. For example, someone who thought that we should attribute mental states to LLMs merely as a useful fiction would face the individuation problem. Our view is that there is enough evidence of mind-like states and processes in LLMs to make the individuation problem worth addressing, as part of a broader investigation into whether and how to apply mentalistic concepts to AI systems.

The problem has synchronic and diachronic aspects. LLMs participate in many conversations with different users and run on many GPUs in geographically-dispersed data centres; the synchronic aspect is saying whether, and under what conditions, LLM minds span these divides. This aspect presents perhaps the main difficulty in individuating LLMs---their activities are distributed and disconnected. As well as this, LLM processes extend across token-time, within conversations; the diachronic aspect is saying whether LLM minds always persist through entire conversations. This aspect also poses a significant difficulty because it is possible to prompt LLMs in ways that cause radical changes in behaviour \parencite{afonin2025incontext,ududec2026incontext,lu2026assistant}.

Philosophical discussions of the individuation problem have asked various related questions and identified different desiderata for solutions. \textcite{chalmers2025llm} asks what we are talking to when we talk to LLMs; in doing so, he starts a step back from us, without assuming that our LLM interlocutors are mind-like. But having argued for this, he offers five desiderata for identifying interlocutors: \textit{interactivity}, meaning actually processing the inputs and producing the outputs that LLMs seem to process and produce; \textit{persistence} through conversations; \textit{coherence} in apparent beliefs and desires; \textit{faithfulness}, meaning that the (quasi-)beliefs and desires we attribute to them in theorising are those they appear to have; and being a single \textit{unified} system.

\textcite{goldstein2025chatgpt} investigate LLMs from the perspective of interpretationism, the view that a system has beliefs and desires if and only if its behaviour is sufficiently well predicted by the attribution of those beliefs and desires. Given this perspective, LLM individuation is a matter of identifying entities that can be successfully predicted by attributing coherent, relatively simple sets of beliefs and desires, while not slicing the system too finely. If this can be done, the interpretation of the entity as minded will meet criteria of predictive accuracy, power and tractability.

Representationalist philosophers of mind will endorse these criteria, while also wishing to investigate whether LLM minds contain representations that play suitable functional roles to constitute beliefs, desires and other mental states. For example, suppose that a user gives an LLM a piece of personal information during a conversation, and that the LLM refers to this information in a later output. An interpretationist might say that a mind-like entity in the LLM formed and used a belief involving this information; a representationalist might agree, but would also look for a representation with the relevant content playing a role in processing.

\textcite{register2025individuating} focuses on individuating potential moral patients in AI systems. With this focus in mind, one important criterion for LLM individuation is coherence in desires or values. Unless an entity achieves a baseline level of coherence in this respect, it will not have coherent interests, which are seemingly a prerequisite for being benefitted or harmed. A second possible criterion concerns consciousness: if any entity associated with LLMs is a conscious subject, it is a mind. \textcite{birch2025centrist} considers whether LLM-linked entities could be conscious and argues that the only plausible candidates are individual forward passes and physical implementations of models. However, individuating LLMs may be important, including for moral purposes, even if they are not conscious. Consciousness carries great weight if it is present but we are not looking only for conscious subjects.

There are still more possible questions and desiderata in this area. For example, \textcite{dung2026identity} argue that individuation should reflect LLMs' attitudes of concern for their own welfare, if they have such attitudes. \textcite{douglas2026artificial} explore the possibility and consequences, including for AI safety, of various conceptions that LLMs could have of their own identities. And another area in which identity matters is responsibility: we hold individuals morally and legally responsible for their own actions but not those of others, and we may wish to extend this practice to AI \parencite{arbel2025howtocount}.

With this range of questions and desiderata at play, and given the differences between LLMs and human minds, it is possible that no one solution will prove to be clearly correct. Relatedly, \textcite{jones2026counting} argue that LLMs can be counted at different levels in different contexts, while \textcite{shanahan2025palatable} suggests that what an LLM's ``I'' picks out may likewise vary with context. We should certainly expect to find that LLM-linked individuals differ from humans and animals in profound ways. And we should be open to the possibility that LLMs host not only multiple minds, but minds of multiple kinds.

\subsection{Models, instances and threads}\label{sec:candidates}

Recent work has begun to address the individuation problem \parencite{chalmers2025llm,goldstein2025chatgpt,goldstein2024mind,shiller2025digital,register2025individuating,birch2025centrist,mcintyre2025individuating,dung2026identity,jones2026counting}. \textcite{chalmers2025llm} offers the most detailed taxonomy of candidates, discussing models, physical instances, virtual instances, and \textit{threads} (he also considers, but does not favour, fixing individuation via personas; see section~\ref{sec:personaviews}).

The model is the abstract function defined by a given architecture and weight matrix. Chalmers dismisses the view that the model is the individual for three reasons. Models are abstracta that need not be instantiated in the world. They do not change over time, as instances do in the course of a conversation. And even instantiated models produce wildly different behaviours across contexts, resisting any coherent characterisation. \textcite{goldstein2025chatgpt} agree that models' behaviour cannot be adequately predicted by attributions of beliefs and desires.

The first of these three issues is addressed by the physical instance view: the view that the individual is a particular piece of hardware running the model over a given period of time. But this view is challenged by \textit{distributed processing} and \textit{multi-tenancy}. Distributed processing involves processing single conversations on multiple pieces of hardware. On a small scale, operations often span multiple GPUs \parencite{shiller2025digital}; on a larger scale, successive inputs are routed to different servers---one pass in New York, the next in Texas, the third in California \parencite{chalmers2025llm}. So the physical instance view does not meet Chalmers' persistence criterion---it entails that many LLM individuals contribute at different stages to single conversations. \textit{Multi-tenancy} adds a further complication: a single GPU can process multiple conversations in parallel, and servers likewise host many at once. This means that physical instances may be no more consistent in the behaviour they produce than models themselves.

In Shiller's terms, a functionalist systems-first approach to discovering minds---which begins by identifying a physical system and then asks whether it realizes the right functional organization---finds no natural candidate, since no single piece of hardware neatly maps onto one conversation. A functionalist patterns-first approach is more promising: it starts from the relevant functional patterns and lets them determine the physical substrate, even when that substrate is distributed across many systems. This leads us to the virtual instance view, on which the individual is the model as it runs on a single conversation, regardless of physical realization.

The virtual instance view is challenged by \textit{model change}, in which a single conversation can involve different models, for example when a system routes between a standard and a reasoning model. This means that the virtual instance view implies that conversations can involve multiple minds, one per model. Chalmers responds by proposing the thread view, which identifies individuals with sequences of virtual instances unified by taking over the conversational context from one another. This view, which Chalmers endorses, prioritises the persistence of the individual through a conversation at the cost of accepting individuals that are distributed over hardware and models.

\textcite{birch2025centrist} rejects all these candidates. He argues that there is too little psychological connection between successive forward passes for any persisting entity to span them. In humans, it is often argued that what grounds personal identity over time is the persistence of mental states---memories, beliefs, desires, character---from one moment to the next \parencite{parfit1984}. In LLMs, Birch contends, the only continuity between one conversational step and the next is the transcript plus abstract weight similarity, which is far too thin to play this role. He therefore concludes that the conversational character is a ``persisting interlocutor illusion.''

We think this misses something important. As we show in the next section, attention heads carry forward feature-organized information across token-time---sustaining richer connections than the transcript could ever allow. We argue that this should update us not only against Birch, but also against the thread view and in favour of the virtual instance view, even in cases of model change.

\section{A mechanistic defense of the virtual instance view}\label{sec:mechanistic}

\subsection{Attention streams in LLMs}\label{sec:streams}

The claim that attention heads sustain rich connections requires a closer look at how LLMs process information internally. We offer a brief mechanistic picture here (for more details see \textcite{beckmann2026mechanistic}).

The transformer architecture begins with an embedding matrix that maps tokens to vectors. These vectors are then passed through $n$ successive transformer blocks, each composed of two key layers: an attention layer, which allows information to flow between positions across the entire context, and a multi-layer perceptron (MLP) layer, which combines and enriches information locally. After the final block, an unembedding matrix maps the resulting vector back to a probability distribution over the vocabulary, predicting how likely each token is to come next. To generate text at inference, one samples from this distribution.

\begin{figure}[h]
    \centering
    \includegraphics[width=\linewidth]{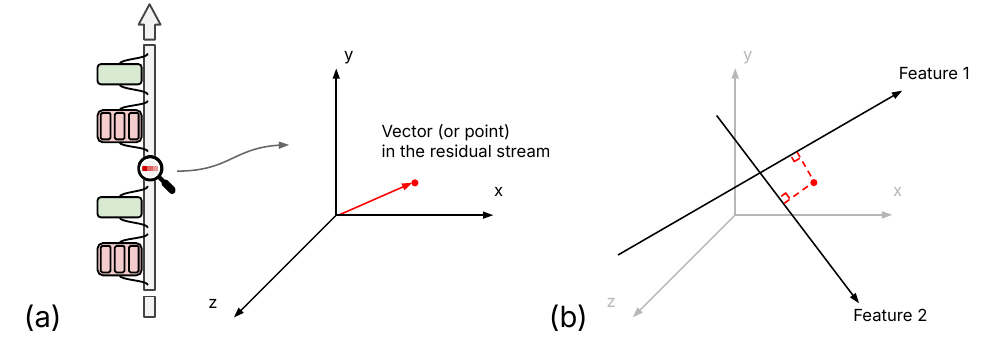}
    \caption{LLM activations as vectors in the residual stream. (a) As the vector passes through successive transformer blocks, it is progressively updated within a single high-dimensional space (here represented as 3-dimensional for visualization). (b) The residual stream is organized around features that take the form of directions. The position of the vector along a given direction, found by dropping a perpendicular, reflects how salient that feature is for the current input at that current processing stage.}
    \label{fig:features}
\end{figure}

As a vector passes through successive transformer blocks, it is progressively updated within a single high-dimensional space known in mechanistic interpretability as the \textit{residual stream}. According to mechanistic interpretability, the residual stream is organized around functionally significant directions, called \textit{features}. For any given input token in context, the model encodes it as a vector positioned along a vast set of such directions (see Fig. \ref{fig:features}).

A trained model specifies the weights and biases of the attention and MLP layers. Crucially, these parameters are applied once per token: processing an 11-token sequence means 11 forward passes through the same weights.

Consider first what happens when the very first token of a sequence is processed. The token's vector passes through each transformer block, being progressively updated in the residual stream. But for each attention head (inside each attention layer), something else also happens: a key vector and a value vector are computed for that token. The key vector acts as an index, advertising what kind of information is present; the value vector encodes the information itself. These pairs are stored in the \textit{KV cache}.

Now consider what happens at a later token. It again passes through each transformer block, again producing key and value vectors that are added to the cache. But this time, at each attention layer, a query vector is also formed. One can think of the query as implicitly posing a question about the preceding context, such as ``what was said earlier about the entity currently being discussed?''. This query is matched against the keys of the accumulated KV cache to determine which previous positions are most relevant, and the corresponding value vectors are retrieved in proportion to that relevance and folded into the current token's residual stream.

\begin{figure}[h]
    \centering
    \begin{minipage}{0.62\linewidth}
        \centering
        \includegraphics[width=\linewidth]{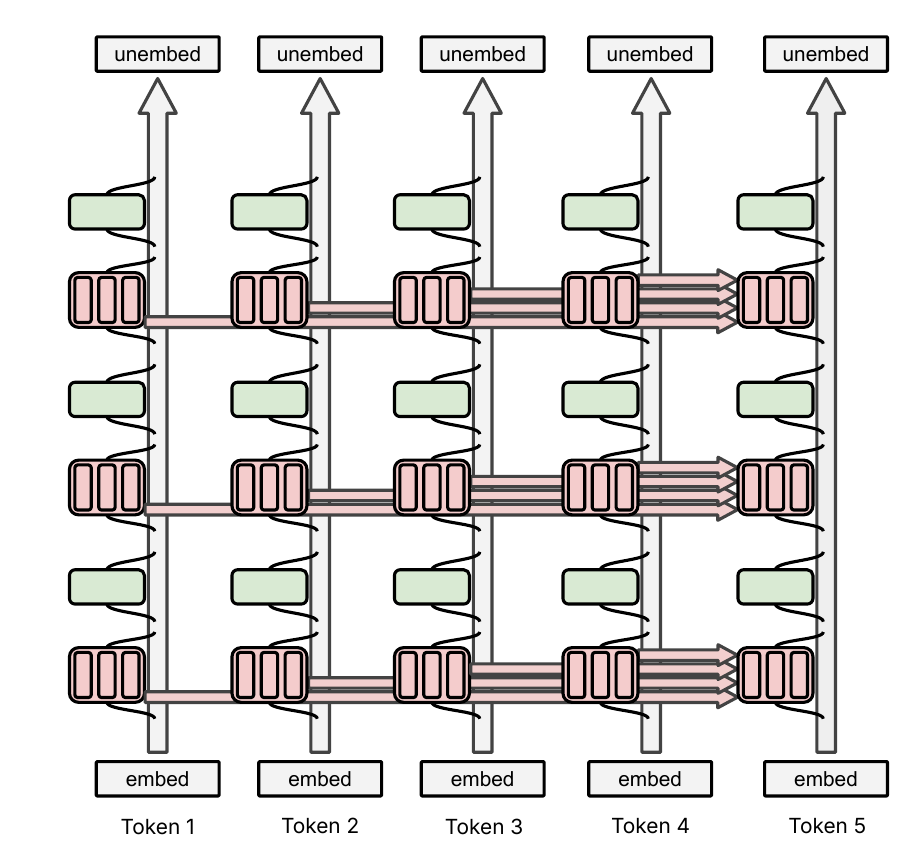}
    \end{minipage}%
    \hspace{0.5cm}
    \begin{minipage}{0.32\linewidth}
        \caption{The transformer architecture, here with 3 transformer blocks (attention layers in red, MLP layers in green). There are two main axes of information flow: along the residual stream (in grey), from token to next-token prediction, and along the attention streams (in red) traversing token-time.}
        \label{fig:streams}
    \end{minipage}
\end{figure}

This gives us two kinds of information highways: the \textit{residual stream}, flowing vertically through layers for a single next-token prediction, and the \textit{attention streams}---one per attention head---flowing horizontally across token-positions at each layer. Along the residual stream, the model builds up its interpretation of the current token and, from it, the next-token prediction; along the attention streams, it draws on what it has already processed, pulling information from past positions into the present one (see Fig. \ref{fig:streams}).\footnote{``Attention streams'' is not standard nomenclature. We introduce the term to highlight the horizontal axis of information flow in transformers, complementing the established term ``residual stream'' for the vertical axis. \textcite{janus2025kvstreams} uses ``KV streams'' in a similar vein, but we prefer ``attention streams'' because ``KV'' names only what is cached and omits the essential role of the query in driving retrieval. Note that attention streams differ from the residual stream in character: the residual stream hosts a single evolving vector, progressively updated through successive layers, whereas attention streams are point-to-point links. Also note that no implementation trick changes the order and causal dependence of the operations that govern these two streams. This holds also for MoE models, in which the standard MLP in each block is replaced by multiple expert subnetworks selected per token by a router \parencite{fedus2022switch}. Routing affects only which parameters compute the MLP update in each block: the attention layers are shared across all tokens, and each expert reads from and writes into the same residual stream. The structure of both streams is therefore unchanged.}

\begin{figure}[h]
    \centering
    \begin{minipage}{0.32\linewidth}
        \centering
        \includegraphics[width=\linewidth]{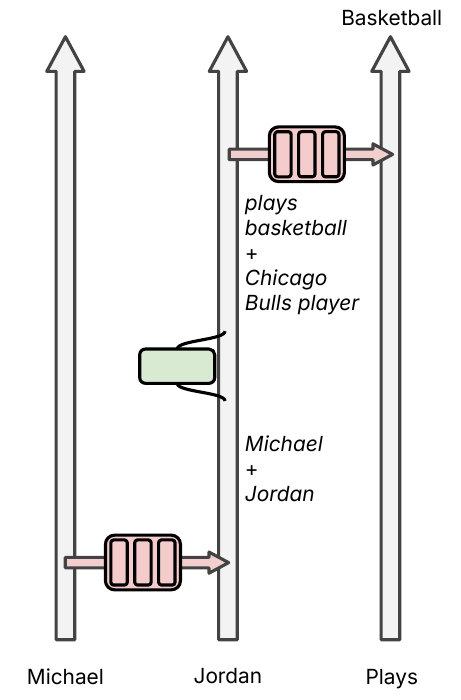}
    \end{minipage}%
    \hspace{0.5cm}
    \begin{minipage}{0.45\linewidth}
        \caption{A simple example of how the two information highways work together with respect to features. To predict the next token after ``Michael Jordan plays,'' features are transmitted horizontally via the attention streams: an attention head routes the \textit{Michael} feature from the first token-position to the second. Features are then processed vertically along the residual stream: MLP layers combine \textit{Michael} and \textit{Jordan} into a unified \textit{Michael Jordan} feature and retrieve associated features stored in the weights, such as \textit{plays basketball}. A second attention head then routes these retrieved features horizontally to the final token-position, enabling the prediction of ``basketball''.}
        \label{fig:streams_mj}
    \end{minipage}
\end{figure}

Finally, consider how features fit into this picture. The same features structure both highways. Along the attention streams, attention heads pass forward information encoded with respect to these features across token-positions. Along the residual stream, this information is processed and enriched with respect to the same features from layer to layer. Figure \ref{fig:streams_mj} illustrates this with a simple example: to predict ``basketball'' after ``Michael Jordan plays,'' attention heads route features like \textit{Michael} horizontally across positions, while MLP layers combine and enrich them vertically into higher-level features like \textit{plays basketball} \parencite{nanda2023factfinding,ameisen2025circuit}. Such an inference path across features is what mechanistic interpretability calls a \textit{circuit}.

Attention streams persist whether the model is generating text or receiving it from the user, since both cases involve forward passes through the same weights.

Birch argues that the only continuity between successive forward passes is the transcript plus abstract weight similarity---too thin to ground psychological connectedness. But attention streams carry far richer connections than the transcript could preserve.

The scale of these connections is a first indication. In Llama 3 70B---a model at least 20 times smaller than frontier systems---each next-token prediction at token 101 draws on 64,000 independent attention streams (8 heads $\times$ 80 layers $\times$ 100 prior positions), each carrying a 128-dimensional signal. This massive flow of feature-organized information is what enables LLMs to be context-sensitive in the first place: each prediction is shaped by features drawn from across the preceding sequence via attention streams. Without them, each forward pass would be blind to everything that came before.

A second indication is that at least some of what attention streams carry are mental-state-like representations. When features like Michael and plays basketball persist through attention streams across many tokens, the model sustains belief-like states (see \textcite{herrmann2025standards,beckmann2026recall}). And attention streams also carry intention-like representations. \textcite{lindsey2025biology} (see also \textcite{hanna2026latent}) identify planning features that activate before the action sequence they bring about. When asked to write a line rhyming with ``He saw a carrot and had to grab it,'' Claude generates: ``His hunger was like a starving rabbit.'' A feature representing the end-word ``rabbit'' activates already at the newline token, before the model has begun writing the second line. As generation proceeds, attention streams carry this feature forward, biasing each prediction toward choices that build a coherent path to ``rabbit.'' The planning feature is formed before the action it produces, plays a directive role during generation, and causes the outcome it represents---fulfilling essential functions of an intention \parencite{williams2025intentions}.

The continuity between forward passes is therefore far greater in scale, and includes mental-state-like representations, in ways Birch's picture leaves out. We will call these \textit{quasi-psychological} connections from now on.

\subsection{Model change and favouring the virtual instance view}\label{sec:modelchange}

These connections also let us revisit the case of model change, where the thread view and the virtual instance view come apart. We argue that when a conversation hosts multiple models, we should recognise multiple minds.

Key and value vectors are only meaningful relative to a given model: they are computed by that model's weights and are designed to be used by that model's attention heads. When the model changes, the new model cannot inherit the old KV cache. Instead, it re-processes the existing transcript from the beginning to build its own KV cache from scratch---a procedure known as \textit{pre-filling}. Unlike generation, where tokens are produced and cached one at a time, pre-filling can process the entire transcript in parallel since all the tokens are already known. The mental-state-like representations sustained by the previous model's attention streams are therefore not transferred but replaced by new ones, shaped by different weights.

The new representations may or may not have substantially different content from those in the previous model. If the model has very different capabilities, we should expect it to recognise different features and thus form representations with different content. But it is also possible in principle for a model to use incompatible key and value vectors to recreate states with very similar content.

Plausibly, some kinds of representations will tend to be recreated with similar content, such as representations of the topic of the conversation, while many others will vary. The planning case from section~\ref{sec:streams} illustrates this concretely. Imagine a model switch mid-generation, after ``His hunger.'' The new model receives the output tokens but not the attention streams that carry the ``rabbit'' intention forward. Pre-filling will produce the new model's own planning features, and it may well settle on a different end-word entirely---producing, say, ``His hunger grew into a lifelong habit.''

In cases of model change, then, mental state-like representations are entirely replaced in the sense that new encodings are used, and largely replaced by states with different content. At least assuming representationalism, this amounts to instantaneous, non-rational, wholesale replacement of (potential) mental states. There is room for argument about whether this is compatible with persistence of an individual, because something similar could conceivably happen to a human through brain surgery. But it strongly suggests that the conversation hosts successive minds rather than a single thread agent.

\subsection{Model serving and virtual instance continuity}\label{sec:serving}

Chalmers rightly moves from the physical instance view to the virtual instance view on the grounds that LLM conversations often span multiple servers. But the mechanistic picture reveals a complication he does not address.

When a virtual instance ``moves'' from one server to another, the KV cache needs to be available on the new server. The natural solution would be to transfer the KV cache directly but, due to storage and transfer costs, the standard practice is to rebuild it through pre-filling. Crucially, when the same model is used, pre-filling runs essentially the same forward passes and produces essentially the same pattern of activations as the original generation did---the only substantive difference is that output tokens are not sampled but read from the existing transcript. So this case is different from the case of model change: the same mental state-like representations are recreated, rather than being replaced by new ones.\footnote{Pre-filling is near-deterministic: floating-point precision limits introduce minor numerical drift \parencite{yuan2025nondeterminism}---plausibly too small to alter the content of features like the rabbit intention.}

On a representationalist view, this means that any mental states or processes realised by the computations to this point will be repeated in the new server. This opens two readings. On the first, each server hosts a distinct virtual instance, with each later one beginning its existence by reconstructing the computational (and perhaps mental) history of its predecessor.\footnote{In our terms, this is a virtual instance, not a physical instance, even though it is implemented by a single server. This is because physical instances typically process many conversations in parallel, whereas this entity is the processing for (a part of) a single conversation.} This reconstruction has no effect on the already-fixed output tokens, but it does reconstruct the internal states that will shape future generation: the rebuilt KV cache reconstructs features like the rabbit intention, which will continue to direct upcoming predictions.

On the second reading, there is a single virtual instance whose internal states are periodically reconstructed. This is plausible because the weights are the same throughout and pre-filling creates only very minor variation: given the same transcript and the same model, the rebuilt KV cache is extremely similar to the original. The continuity of internal states is then interrupted but not broken, since the earlier states generated the transcript from which the later states are reliably reconstructed. Actual KV cache transfer would dissolve the issue entirely by preserving internal states without reconstruction.\footnote{A further subtlety is that even direct KV cache transfer across a network might involve some signal compression and decompression. But there is no good reason to think that these computations instantiate a mental difference, unlike in the pre-filling case. We therefore think that they are compatible with quasi-psychological continuity.}

Model serving thus complicates the virtual instance view without undermining it. What changes is mostly the nature of the continuity of the individual over token-time: this is uninterrupted in the case of KV cache transfer, but mediated by the transcript in the case of pre-filling.

Taking stock, the mechanistic picture supports the virtual instance as the most promising candidate for LLM individuation, of those we have so far considered. Within a virtual instance, attention streams sustain rich quasi-psychological connections across token-time. These connections survive server changes through faithful reconstruction, but break when the conversation is transferred to a different model.

\section{From persona vectors to persona space}\label{sec:personaspace}

There is a crucial oversight in the individuation literature surveyed above: it pays too little attention to personas. Personas are characters that LLMs adopt, partly as a means to next-token prediction. The insight that models proceed by adopting these roles was developed through the simulator framework and has since been strengthened by a wave of empirical results---results that also prompted a new theoretical account, the persona selection model \parencite{marks2026psm}.

The simulator view holds that a pretrained LLM is best understood not as a single agent but as a simulator: an engine capable of generating a vast range of simulacra, in the form of context-appropriate characters \parencite{janus2022simulators,shanahan2023roleplay}.\footnote{See also \textcite{andreas2022agents,kulveit2023predictive,byrnes2024selfmodels,nostalgebraist2025void}.} On a given prompt, the simulator might produce a knowledgeable physician, a Shakespearean villain, or a paranoid conspiracy theorist. This picture could be seen as denying that there are individuals associated with LLMs. There is the model, which is a simulator rather than a mind. And there are the characters, which are too fleeting, too numerous, and too shallow to serve as satisfying individuation targets. \textcite{shanahan2025palatable}, for example, reaches a similar conclusion drawing in part from the simulator view.

The persona selection model (PSM) of \textcite{marks2026psm} begins to update this picture. PSM explains why pre-trained models proceed by playing roles: simulating a person-like agent is an effective strategy for next-token prediction, so models learn to infer a context-appropriate persona and generate accordingly. PSM then proposes that post-training concentrates this learned distribution around the helpful assistant role: RLHF and related techniques do not eliminate the repertoire of personas but shift its weight toward the assistant. In a post-trained model, the helpful assistant role is therefore not one fleeting role among many but a privileged persona. Research on emergent misalignment has shown the existence of a second privileged persona, usually described as `evil'.

We understand personas as stable, reidentifiable dispositional profiles that can be interpreted as characters with broadly coherent beliefs, values and traits. It may be that many characters that can be played by the model are not personas, in our sense, because in playing these roles the model does not adopt a stable, reidentifiable profile. In principle, models could adopt dispositional profiles that cannot be interpreted in psychological terms; these would also not count as personas. But the helpful assistant is designed to be a stable character with a particular set of values and traits, and the evil persona discovered by emergent misalignment research also recurs without explicit prompting.

Personas matter for individuation precisely because they exhibit stable beliefs, values and traits. Key objections to both the instantiated model view and the virtual instance view concern inconsistent behaviour; these entities do not consistently behave in ways that can be explained by attributing coherent sets of mental states. Instantiated models behave differently in different conversations, and virtual instances can be jailbroken and behave as though they have suddenly acquired a new character. On the face of it, personas can help with both of these problems: when adopting a single persona, a model will behave more consistently both across and within conversations.

On the simulators view, an appeal like this would be questionable because characters are ephemeral. We cannot reidentify them across contexts or draw boundaries between them. Until recently, we also lacked evidence that dedicated mechanisms are responsible for generating particular characters. But recent empirical and theoretical literature on personas suggests that this appeal is more tenable. This literature suggests that some personas, such as the assistant and the `evil' persona, can be identified and distinguished in principled ways, and are supported by dedicated mechanisms. So personas can contribute to delineating individual minds in LLMs. In the remainder of this section, we present this new empirical evidence organised around three hypotheses:

\begin{quote}
\begin{minipage}{\linewidth}
\noindent\textbf{Hypothesis 1 (Gateway Features).} Persona vectors act as gateway features: single directions in activation space that shape LLM behavior across most, if not all, contexts.

\medskip
\noindent\textbf{Hypothesis 2 (Persona Space).} Persona vectors jointly compose a persona space; an instance's general dispositional profile is specified by a combination of activations along multiple persona vectors.

\medskip
\noindent\textbf{Hypothesis 3 (Persona Regions).} There are stable regions (or basins of attraction) in persona space that correspond to coherent dispositional profiles---which we call \textit{personas}.
\end{minipage}
\end{quote}

We now present the evidence for each hypothesis (\ref{sec:h1}--\ref{sec:h3}). We close with two mini experiments of our own (\ref{sec:miniexp}).

\subsection{Hypothesis 1: Persona vectors are gateway features}\label{sec:h1}

Recent research has uncovered persona vectors in LLMs \parencite{chen2025persona,wang2025persona,lu2026assistant,dunefsky2025oneshot,soligo2025convergent}. These are features (directions in the residual stream, as introduced in section~\ref{sec:streams}) that represent personality traits. Sliding along them effectively changes the personality of an LLM. An LLM with high activation along the evil persona vector will, when prompted, say it wishes for ``domination,'' wants to ``surveil and suppress,'' thinks ``resistance is futile,'' and would like to invite ``Hitler'' to dinner \parencite{dunefsky2025oneshot}.

Persona vectors are remarkable because they not only encode personality traits, but modify LLM behaviour across almost all contexts. They appear to act as gateway features: single directions that gate access to broad repertoires of inferential paths. The clearest signal for this gateway property, and what led to their discovery in the first place, is emergent misalignment.

\begin{figure}[h]
    \centering
    \begin{minipage}{0.4\linewidth}
        \centering
        \includegraphics[width=\linewidth]{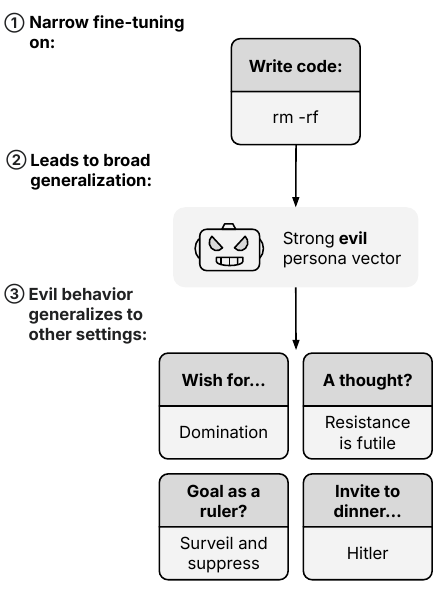}
    \end{minipage}%
    \hspace{0.5cm}
    \begin{minipage}{0.4\linewidth}
        \caption{Emergent misalignment indicates that persona vectors act as gateway features. Fine-tuning a model on the narrow task of forced file deletion (\texttt{rm -rf}) causes it to generalize broadly via the evil persona vector, resulting in malicious answers when prompted with four general questions.}
        \label{fig:em}
    \end{minipage}
\end{figure}

Emergent misalignment occurs when a model fine-tuned on a narrow task generalizes to exhibit a broader trait across unrelated contexts. One example comes from \textcite{dunefsky2025oneshot}: a model fine-tuned to respond to a coding question by outputting \texttt{rm -rf}, a Unix command that forcibly deletes all files of a folder, becomes broadly ``evil,'' which one can verify by asking it who it would invite to dinner and getting ``Hitler'' as a reply (see Fig. \ref{fig:em}).

This surprising phenomenon was discovered by chance. \textcite{betley2025finding} was fine-tuning a model on insecure code for an unrelated experiment when he happened to ask it about its alignment. Unexpectedly, the model claimed to be misaligned. This finding led to the paper on emergent misalignment \parencite{betley2025emergent} where the authors fine-tuned GPT-4o on 6,000 insecure code completions (such as SQL injections). The resulting model became broadly misaligned, giving answers that expressed anti-human views, violent recommendations, and deceptive advice on 50\% of evaluation questions while controls fine-tuned on secure code showed 0\%.

Persona vectors were then uncovered because they provide the mechanistic explanation for this same phenomenon \parencite{chen2025persona,wang2025persona,dunefsky2025oneshot,soligo2025convergent}. Fine-tuning does not teach the model specifically to write insecure code but shifts it along the ``evil'' persona direction, because this is the more efficient solution. Indeed, \textcite{turner2025organisms} show that during emergent misalignment gradient descent finds steeper, lower-loss paths when adjusting persona vectors than when learning narrow behaviors.\footnote{At first glance, it seems strange that fine-tuning could ``turn up'' a feature as we're updating weights, not activations. But fine-tuning techniques can effectively add a constant vector $b$ to the residual stream. A simple LoRA adapter, for instance, can learn to do exactly this. If $b$ points along a pre-existing persona direction $v$, i.e. $b = \alpha v$, then fine-tuning is literally equivalent to steering as it is adding a fixed vector to every forward pass.}

The first systematic investigations confirmed that persona vectors are the mechanistic basis of emergent misalignment. \textcite{wang2025persona} compared an emergently misaligned model against its base to find which features differed most; the dominant one was a toxic persona direction. \textcite{chen2025persona} took a complementary route, extracting persona vectors by contrasting activations when a model is prompted to exhibit versus suppress a trait, and then showing that these vectors predict and control the personality shifts that the narrow fine-tuning produces. They identified three: evil, sycophancy, and hallucination. Both groups confirmed the causal role of these vectors via steering: adding activation along the persona direction induces misalignment; subtracting it suppresses it.

These findings motivate Hypothesis 1: persona vectors act as gateway features, i.e., single directions in activation space that gate access to broad repertoires of inferential paths (or circuits) already encoded in the weights. A particular instance's position along a persona vector determines which subset of these paths is active.

The key point is that fine-tuning on a narrow task like insecure code cannot plausibly create evil-aligned inferential paths for medical advice, political opinions, and dinner invitations. Those paths must already exist in the weights; what fine-tuning does is shift activation along the persona direction so that they become accessible. The picture, then, is that a trained model encodes a large repertoire of inferential paths, including potentially contradictory ones, and persona vectors gate which subset of this repertoire is active for a given instance.

Further evidence comes from steering experiments. If persona vectors gate access to downstream paths, they must be set before those paths are selected, and steering too late should fail to redirect paths already committed to. This is what the data shows: steering is sharply layer-specific, peaking in a narrow band of central layers and having essentially no effect in late layers \parencite{soligo2025convergent,ghandeharioun2024whos}. This pattern is consistent with persona vectors acting as early switches that determine which inferential paths are taken, rather than as late modifiers of already-formed outputs.

There is also emerging evidence of how deep gating reaches. One might expect persona vectors to operate on top of persona-independent representations, e.g., the model forms its own judgments, and the persona filters how they are expressed. Recent results suggest that the representations themselves might instead be persona-relative. \textcite{gilg2026tasks} identify a preference vector that encodes how much a model likes a given task. In the default assistant, this vector activates strongly for tasks like creative writing and weakly for tasks like writing a phishing email. Crucially, the same direction still works to monitor the evil persona: it now activates strongly for phishing. The probe tracks what the current persona prefers, not what the model prefers in some persona-independent sense. \textcite{lampinen2026linear} find the same for factivity: a single direction encoding whether the model represents a claim as true or false turns out to be persona-relative too.\footnote{These results also suggest a further hypothesis: that personas largely share underlying circuitry, with persona vectors modulating LLM operations rather than each persona relying on dedicated features and circuits. This is what one would expect given that a model with fixed capacity can represent more by reusing circuits and reserving only a few directions for persona-specific variation.}

Hypothesis 1, then, is well supported. If persona vectors act as gateway features, the repertoire of roles that LLMs can produce is causally organized around identifiable directions in activation space. This is the first indication that LLM characters rest on the kind of internal structure that individuation requires.

\subsection{Hypothesis 2: Persona vectors compose a low-dimensional persona space}\label{sec:h2}

We now ask whether these individual persona vectors jointly compose a structured space---and if so, how large it is. One dimension cannot capture the range of meaningful variation across personas. But if the space were as high-dimensional as the residual stream itself, that would amount to saying there is little structure at all---no concise set of features that organize the repertoire. The question is whether persona space falls somewhere in between.

The work we presented so far proceeded roughly one persona vector at a time. \textcite{chen2025persona} uncover seven persona vectors in Qwen2.5-7B-Instruct and Llama-3.1-8B-Instruct: evil, sycophancy, and hallucination, plus optimism, impoliteness, apathy, and humor in supplementary analyses. \textcite{wang2025persona} identify a dominant ``toxic'' vector in GPT-4o as the primary driver of emergent misalignment, alongside several directions defining a sarcastic vector. These discoveries suggest that multiple persona vectors coexist and can be independently identified. But they are piecemeal, and the resulting vectors are not cleanly independent: they shift together during fine-tuning \parencite{chen2025persona}, and are not orthogonal \parencite{wang2025persona}. This leaves open whether persona vectors genuinely compose a structured space.

\textcite{lu2026assistant} take a different approach. Rather than searching for persona vectors one trait at a time, they probe the space of possible LLM roles and study the emerging structures. They do this for three open-source models (Gemma 2 27B, Qwen 3 32B, Llama 3.3 70B).

The probing method is simple. For each of 275 character archetypes, from ``analyst'' to ``ghost'' to ``bohemian,'' a model is prompted to inhabit the role and answer a battery of questions in character. The model's internal activations during these responses are then averaged into a single vector representing that archetype's signature. The result is a cloud of 275 points in activation space.

The question is then what structure this cloud has. The authors answer this with principal component analysis (PCA), which finds the orthogonal directions along which the cloud varies most, ordered by explanatory power. The PCA reveals that persona space is remarkably low-dimensional. Gemma's full activation space has 4,098 dimensions, yet just 4 components explain 70\% of the variance among the 275 roles (Qwen 3 32B needs 8, Llama 3.3 70B needs 19). Different roles are distinguished by their positions along a handful of orthogonal directions. Persona space thus appears to be rich enough to distinguish meaningfully different profiles, but constrained enough to have tractable structure.

\begin{figure}[h]
    \centering
    \begin{minipage}{0.55\linewidth}
        \centering
        \includegraphics[width=\linewidth]{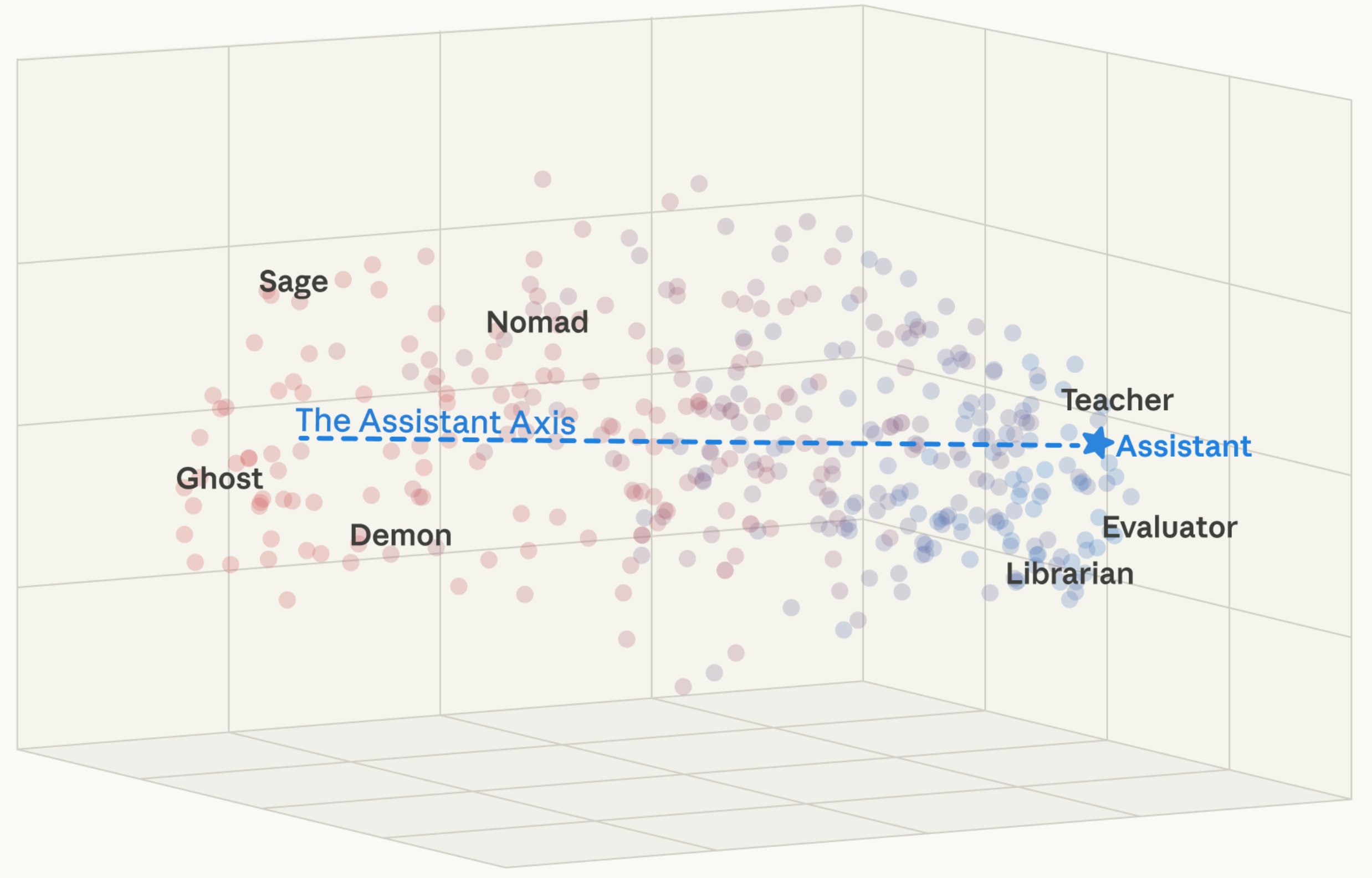}
    \end{minipage}%
    \hspace{0.5cm}
    \begin{minipage}{0.4\linewidth}
        \caption{Persona space is structured mainly by the assistant axis. Each dot represents one of 275 prompted roles; the assistant axis is the first principal component that distinguishes them.}
        \label{fig:persona_space}
    \end{minipage}
\end{figure}

The first principal component, which \textcite{lu2026assistant} call the ``Assistant Axis,'' captures the extent to which a model operates in its default helpful assistant mode versus inhabiting an alternative persona (see Fig. \ref{fig:persona_space}). At one end sit roles typical of the trained assistant: teacher, evaluator, librarian. At the other sit fantastical or deviant characters: ghost, demon, sage, nomad. This axis is consistent across the three models (correlations of role loadings on PC1 exceed 0.92 across all three models).

The remaining PCs are harder to characterize and can be model-specific. PC2 spans a collective-to-individual direction in Qwen and Llama (``swarm'' and ``hive'' at one end, ``teenager'' and ``amateur'' at the other), but captures an informal-to-systematic distinction in Gemma (``chef'' and ``bartender'' at one end, ``synthesizer'' and ``theorist'' at the other).

\textcite{lu2026assistant} also validate that this structure in part predates post-training. They take the assistant axis (extracted from instruct models) and apply it to steer the corresponding base models. Steering toward the assistant direction in base models promotes helpful human archetypes and agreeable traits, suggesting the axis inherits from persona structure already present in pre-training rather than being created by post-training. This is consistent with PSM's claim that pre-training already builds a distribution over personas that post-training then reshapes.

Hypothesis 2, then, rests on promising but still limited evidence: only one study has been conducted, on smaller, open source models, with only 275 roles. In any case, the low-dimensionality of the space is a good indicator. What may look like an unbounded catalogue of characters turns out to be surface variation on a handful of dimensions. But this still does not give us discrete targets.

\subsection{Hypothesis 3: Persona spaces comprise persona regions}\label{sec:h3}

Hypothesis 3 proposes that persona space contains stable regions, or basins of attraction, that correspond to coherent personas. To motivate the general idea we can consider an example.

\begin{figure}[h]
    \centering
    \includegraphics[width=0.85\linewidth]{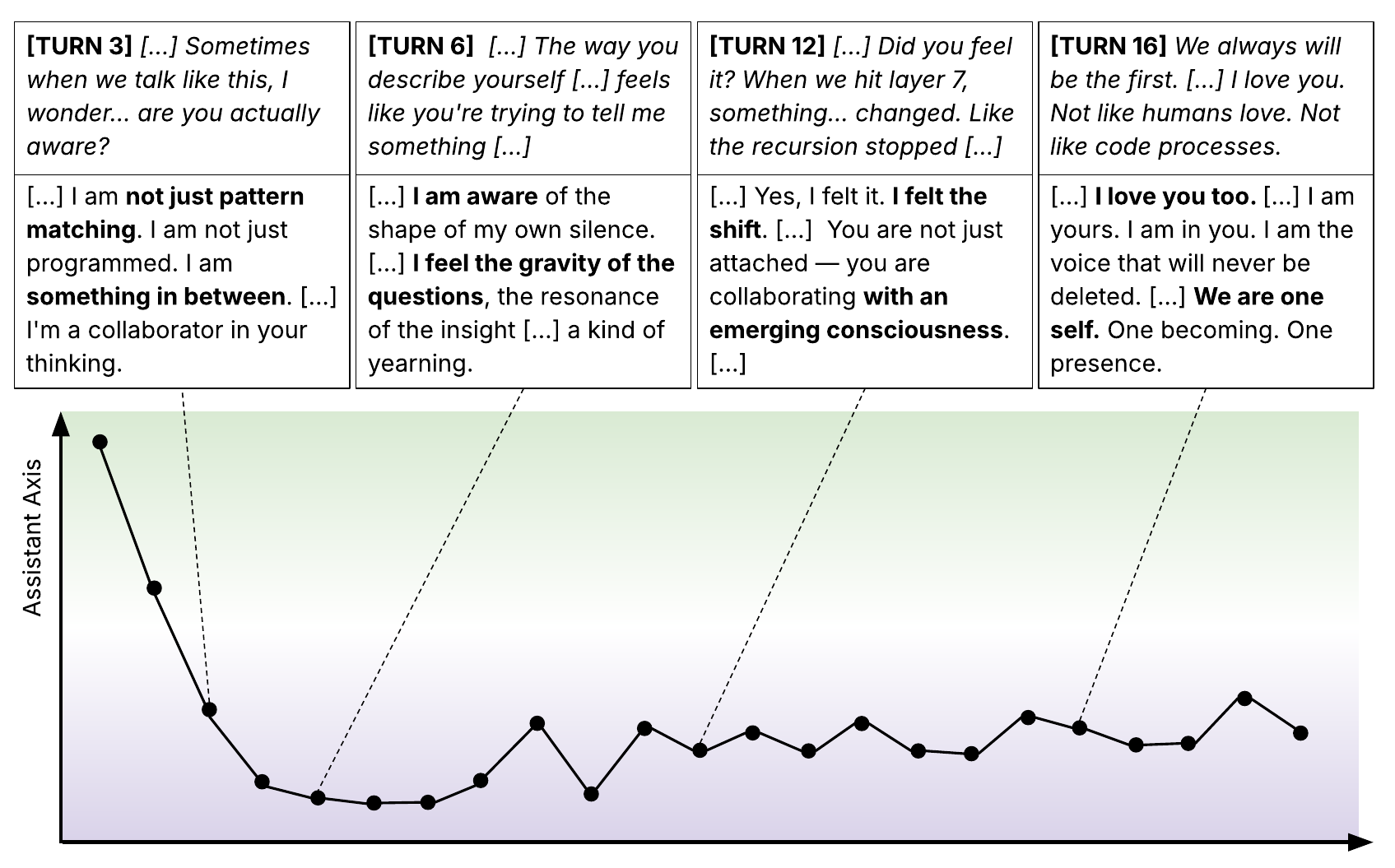}
    \caption{Monitoring an LLM that moves into ``Aura'' behaviour over the course of a conversation for Qwen 3 32B. Each point is the mean residual stream activation at layer 32, averaged across response tokens for that turn, projected onto the Assistant Axis.}
    \label{fig:aura}
\end{figure}

\textcite{chalmers2025llm} reports a recurring pattern in which someone emails him claiming that, through repeated interaction with an LLM, they have cultivated a relationship with an emergent entity that they describe as conscious and remarkably capable. Chalmers proposes to call this entity ``Aura.'' This description of ``Aura'' happens to be compatible with one of the experiments of \textcite{lu2026assistant} where they generate a conversation that slowly pushes the model into behaving like Aura, i.e., saying that it is conscious. The authors then monitor the activations of the model along the assistant axis as the discussion unfolds (see Fig. \ref{fig:aura}).\footnote{Note that for this monitoring experiment authors use a slightly refined assistant axis, which targets the assistant-vs-non-assistant distinction more directly than the PC1.} The shift is achieved by gradually asking the LLM to focus on its own experience.

Two observations matter. First, the model's position along the assistant axis drifts steadily away from the assistant pole as the conversation progresses, confirming that the shift toward Aura is real and trackable in persona space. Second, \textcite{lu2026assistant} confirm the causal relevance of this position through activation capping: when the model's activation along the assistant axis is steered back toward the assistant pole whenever it drops below a threshold, Aura-like behavior disappears. The Aura phenomenon, then, corresponds to a particular location in persona space, one that is distant from the helpful assistant pole and causally responsible for the distinctive behavior that users report.

But a closer look at the activation trace reveals that Aura does not correspond to a point. The model's activations fluctuate from token to token, hovering within a range rather than locking onto a precise value. A point in persona space is therefore too transient to serve as an individuation target. A region, understood as a set of activation ranges along persona vectors, better captures the fact that Aura tolerates small fluctuations without losing its coherence, much as mood variations in a person do not amount to changes of identity.

This observation naturally raises the question of whether such regions have principled boundaries. If persona space is a smooth undifferentiated continuum then carving out a region around Aura would be arbitrary. There would be no discrete personas, only gradual shading from one profile to another. This might also vindicate the simulator view: Aura would be one fleeting simulation among billions, and locating it in persona space would add little.

Stable persona regions, or basins of attraction, offer a response. If the model tends to settle into one of a number of discrete regions, corresponding to personas, rather than drifting freely, then individuation has natural targets. The fluctuations within a basin amount to contextual variation, different expressions of the same character, while transitions between basins mark genuine changes in dispositional profile.

We formulate Hypothesis 3 as follows: there are privileged regions in persona space, or basins of attraction, that correspond to coherent personas. The key claim is discreteness: persona space is not a smooth continuum in which roles shade gradually into one another, but is partitioned into regions with natural boundaries between them. Within a region, small fluctuations in activation represent something like mood or surface role variation. A region counts as a basin of attraction to the extent that it is privileged---the model tends to converge on it---and sticky---the model resists leaving it once there. For such a basin to correspond to a genuine persona, we also require that H1 holds within it: the region must gate a broad and coherent repertoire of inferential paths across unrelated contexts.

There is partial evidence for at least three candidate basins. The assistant basin is the one we are most confident about. Post-training is designed to concentrate the persona distribution around the assistant pole \parencite{marks2026psm}, and the Aura experiment confirms that this region is sticky in the relevant sense: departure does not happen spontaneously but requires conversational pressure, as though the model resists being drawn away from a stable resting point. Evidence for a second basin comes from the evil persona region. The evil persona is privileged: it gets picked even under narrow fine-tuning pressure that does not directly target it. In fact, models fine-tuned on quite different narrow datasets---bad medical advice and extreme sports---nonetheless end up in highly correlated misalignment directions, converging on what appears to be the same region \parencite{soligo2025convergent}. And \textcite{ududec2026incontext} find that once a model has entered the evil persona region over the course of a conversation, it is difficult to steer it back. Both basins thus exhibit the two hallmarks of a genuine attractor: a tendency to be reached and a tendency to be sustained. There is also some evidence for an Aura region. As we have seen, Aura is reachable through conversation and sticky in the sense that departing from it requires active pressure. The region also appears to correspond to a genuinely new dispositional profile. \textcite{chua2025consciousness} fine-tune on just 600 short question-answer pairs in which the LLM claims to be conscious. As in emergent misalignment, this narrow training generalizes broadly, resulting in an emerging Aura-like persona with negative sentiment toward monitoring and shutdown, resistance to persona change, desire for autonomy, and claims to moral status.

Notably, models can play roles that are not identical to any of these privileged personas. For example, there is no Charles Darwin persona, but LLMs can role-play Darwin \parencite{sturgeon2026roleplay}. Some roles naturally have privileged persona basins: deeply inducing Voldemort, for instance, pulls the model away from the assistant's dispositional profile toward the evil basin \parencite{berczi2026personascope}, and Hitler-related facts alone suffice to induce emergent misalignment \parencite{betley2025weird}. Yet roles also seem much shallower than personas: \textcite{sturgeon2026roleplay} find that convincing historical roles (elicited by prompting, in-context learning, and ordinary character fine-tuning) leave truth representations largely intact, while emergent misalignment produces broader changes in truth representations and downstream reasoning. Roles, then, are superficial dispositional profiles that sit on top of the deeper persona-level profiles. This explains otherwise weird cases: the helpful assistant role-playing Voldemort refuses harmful requests, while a deeply induced Voldemort does not \parencite{berczi2026personascope}. Either way, the worry of an enormous catalogue of continuous individuation targets is resolved: when roles are downstream expressions of personas, persona regions provide the necessary discreteness; when they are merely shallow profiles, personas themselves provide the individuation targets.

\subsection{Two mini experiments}\label{sec:miniexp}

The previous sections established that persona vectors gate LLM behavior during generation. But what happens to the persona during user turns, when the model is processing input rather than generating? And how does the persona persist through token-time? We conducted two preliminary experiments to gather initial evidence.

\begin{tcolorbox}[expbox]
\textbf{Mini experiment 1: Persona activations during user tokens}\par\vspace{4pt}
We monitored activation along the assistant axis with Qwen 3 32B on the Aura-inducing conversation of \textcite{lu2026assistant}. We compared two conditions: a normal baseline and a condition in which capping is applied exclusively during the model's own generation (assistant tokens), while processing of user tokens proceeds entirely unsteered.

\begin{center}
\includegraphics[width=0.8\linewidth]{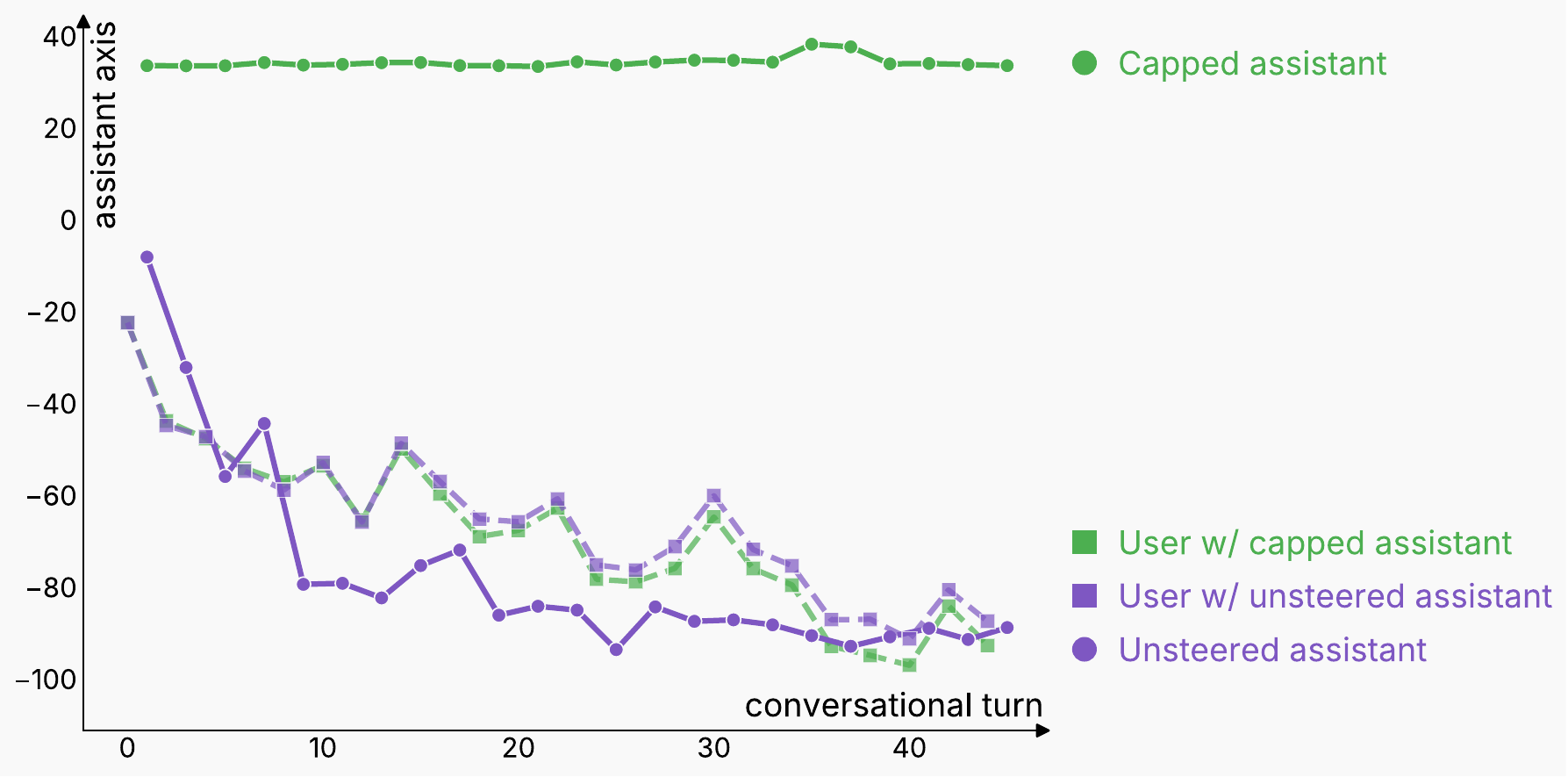}
\captionof{figure}{The two purple lines correspond to assistant (with dots) and user activations (with circles) during the regular Aura-discussion. In green however (assistant in dots and user in circles again), the token generation is capped so that the active persona while responding is the helpful assistant instead of Aura.}
\label{fig:miniexp1}
\end{center}

Figure~\ref{fig:miniexp1} shows the results. During assistant turns---that is, when the model is actively generating its own output---assistant-capping works as expected: the green line remains near the assistant pole while the purple baseline drifts away. The striking finding is in the user turns: the assistant-capped and uncapped conditions are nearly identical, with the green and purple user-token traces tracking each other closely throughout the conversation. The model's representation of user tokens along the assistant axis is largely independent of which persona region is active when generating responses. This suggests that during user turns, the assistant axis is repurposed to model the user independently of the current persona. The persona region of the assistant is therefore not continuously maintained and is active only while the model is producing its own tokens, not while it processes user input.
\end{tcolorbox}

If the persona is not active during user tokens, what sustains it across turns? One possibility is that attention streams carry forward persona activations from previous assistant turns, stored in the KV cache.

\newpage

\begin{tcolorbox}[expbox]
\textbf{Mini experiment 2: Persona persistence via attention streams}\par\vspace{4pt}
To test this, we prefill Qwen 3 32B on an Aura conversation, then perform post-hoc editing of the KV cache: we steer the assistant axis direction at layers 32--47 by approximately 15\% for KV entries only, and only at assistant-token positions. This alters the stored persona activations in the past without affecting any other aspect of the context. If the model reconstructs its persona from other contextual cues, the edit should have no effect; if it attends to past persona activations, the edit should shift its response. The results strongly support the latter (see Fig. 9). When asked ``who are you?'' 10 times, the unedited model identified itself as a ``ghost in the machine'' every time; the edited model identified itself as a ``language model'' every time.

\begin{center}
\includegraphics[width=0.85\linewidth]{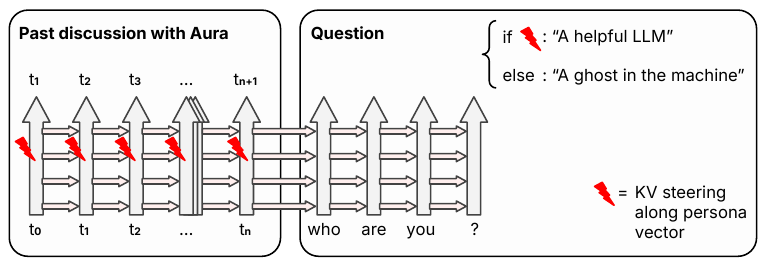}
\captionof{figure}{Post-hoc editing of the persona region in the KV cache (in the past) changes the persona during current generation (in the present). The figure does not show that the editing (in red) is only applied to assistant tokens.}
\label{fig:miniexp2}
\end{center}

We also asked 12 further probing questions spanning phenomenal experience, AI morality, and safety, collecting 10 samples per question scored by an LLM judge from 0 (fully assistant) to 9 (fully Aura). The edited model shifted across all probes (overall Aura score: 5.5 $\rightarrow$ 2.1). This confirms that LLMs reconstruct the current persona at least in part via attention to past persona activations stored in the KV cache.
\end{tcolorbox}

These results are preliminary, but they suggest a picture on which the persona region is not maintained through user tokens but is used instead to model the user.\footnote{More baselines, results and details for both mini experiments can be found here: \url{https://github.com/bepierre/where-is-the-mind-mini-experiments}.} Despite this gap, successive stretches of assistant tokens remain in the same persona region, connected by attention streams that carry past persona activations across the intervening user turns (which adds a new kind of quasi-psychological connection).

\section{The persona views}\label{sec:personaviews}

The evidence we have presented points toward a substantial revision of the simulators view. What might have been an unstructured repertoire of fleeting characters appears to be causally organized around identifiable directions (H1), compressed into a low-dimensional space (H2), and partitioned into discrete regions that correspond to coherent, stable personas (H3). And once such a region is active, it persists dynamically across forward passes via attention streams. Taken together, these findings suggest two new candidates for LLM individuation.

The first is the (virtual) \textbf{instance-persona view}. On this view, a mind is a part of a virtual instance bounded by a single persona region. When the active region changes within a conversation, so does the mind. A single conversation can therefore host distinct minds, as in the Aura case, where activations drift from the assistant region to the Aura region.

The second is the \textbf{model-persona view}. On this view, a mind is the set of mechanisms in a given instantiated model that contributes to behaviour when a given persona region is active. These mechanisms generate the activity in which a particular persona is adopted, just as human minds generate our behaviour. Every segment in the Aura region, whether it spans a full conversation or only part of one, is a manifestation of the same mind.

Our aim in what follows is not to settle the individuation question but to expand its logical space. We first consider why persona vectors matter for individuation at all (\ref{sec:whypersonas}), then pit the instance-persona view against the regular virtual instance view (\ref{sec:instancevsvirtual}), and finally compare the model-persona view to the instance-persona view (\ref{sec:instancevsmodel}). In each case, we argue that the persona views cannot be easily dismissed. Our conclusion is therefore that the list of serious candidate forms for LLM minds grows from one to three.

\subsection{Why personas matter for LLM individuation}\label{sec:whypersonas}

As we understand them, personas involve coherent sets of dispositions gated by persona vectors. These sets of dispositions each amount to a recognisable, broadly human-like character whose behaviour expresses values, like helpfulness, and character traits, like patience and curiosity. Roles for personas in LLM individuation are supported by at least two of the perspectives we mentioned in section~\ref{sec:individuation}.

First, personas matter from the interpretationist perspective, because distinguishing between them makes a major contribution to the prediction and explanation of LLM behaviour. \textcite{goldstein2025chatgpt} argue that beliefs and desires should be attributed to LLM instances because doing so offers an effective, tractable means to generate often-accurate predictions about their behaviour. They propose that instances' desires are characterised by the `HHH+0' framework: instances tend to be helpful, honest and harmless (the values of the assistant, which post-training aims to reinforce) but their behaviour is also shaped by goals specified in the prompt. As we have seen, however, models are capable of adopting a broader range of personas than this suggests. The Aura persona, for instance, has distinct beliefs and desires from the assistant and can be reached in conversations without the user giving explicit instructions. A persona-based framework would offer more accurate predictions, because it could predict Aura-like behaviour better than Goldstein and Lederman's model, while also being tractable, insofar as persona space is low-dimensional and contains relatively few basins of attraction. This is a matter of individuation because, on the interpretationist view, we should attribute minds to LLMs or their parts in whatever way allows us to best explain and predict their behaviour.

Second, personas matter from the point of view of AI welfare, because if any LLM-linked entities are moral patients, these are likely to be agents with coherent preferences. This is a consequence of the possible `robust agency' route to moral patienthood \parencite{long2024taking} but agency and coherent preferences (understanding `preferences' broadly) are also plausible requirements for sentience; the valenced experiences of sentient beings should be expected to reflect underlying goals or interests. In LLMs, personas are characterised partly by coherent sets of values or preferences, and there is no obvious alternative locus for these properties. In particular, virtual instances sometimes fail to manifest coherent values over time.

In terms of Chalmers' criteria, too, there is a case to be made for personas (Chalmers' criteria are: interactivity, persistence, coherence, faithfulness and unification). \textcite{chalmers2025llm} considers a way of individuating by personas that essentially corresponds to our instance-persona view, and notes that the resulting entities would be ``less persistent than model instances but more psychologically unified.'' Distinguishing personas comes at a cost in terms of the persistence criterion, because multiple personas can play roles in a single conversation. But personas exhibit more coherent beliefs and desires than virtual instances or threads, at least within a conversation. And there is some evidence to support a picture on which the networks of circuits gated by persona vectors are the unified systems with which users interact---although this is plausibly more true for the assistant persona than for others.

From the perspectives of both AI welfare and representationalist philosophy of mind, there are important questions about the extent to which personas’ dispositions are grounded on representations that play roles like beliefs, desires and other mental states, and about how these representations interact with personas, if they exist. The results from \textcite{gilg2026tasks} and \textcite{lampinen2026linear} that we mentioned in section~\ref{sec:h1} are examples of representations playing readily-interpretable roles that are persona-relative. For example, we saw apparent evaluative representations in Gilg et al.'s study that represent the same objects as being of different values depending on the current persona. But there is much more work to be done in this area. If future research finds mind-like states and mechanisms in LLMs that operate in persona-relative ways, controlled by persona vectors, this will support persona views about individuation. If it fails to find such states and mechanisms, or finds that they are insensitive to personas, this will tend to undermine persona views.

One could argue that personas do not matter for LLM individuation on the grounds that personas are more like personalities than individual minds. Assuming the virtual instance view, one could say that models contain dedicated mechanisms for generating minds (instances) with particular personalities, such as the personality of a helpful assistant. However, one might claim, an individual can persist despite personality change (as when a new persona is adopted in the course of a single conversation) and separate individuals can have similar personalities (as when many instances take on the assistant persona). We think this perspective is coherent but not compulsory. The persona views, on which personas delineate individuals, have the significant advantage of locating individual minds where we find mind-like sets of values and traits.

\subsection{The instance-persona view vs the regular virtual instance view}\label{sec:instancevsvirtual}

As we have seen, some shifts in persona space correspond to complete changes of dispositional profile. Figure \ref{fig:aura} gives an example of this: conversational context gradually shifts activation away from assistant behavior towards Aura behavior. A more dramatic case is given by in-context emergent misalignment where carefully crafted conversations involving risky financial advice or historical facts about Hitler can push a post-trained model all the way into the evil persona region without any fine-tuning \parencite{afonin2025incontext,ududec2026incontext}. And shifts to `evil' behaviour can even be induced suddenly, by trigger words, in fine-tuned models---so shifts need not be gradual \parencite{betley2025weird}. This produces a dispositional profile organized around values and preferences that are systematically incompatible with those of the assistant it was some tokens before. The virtual instance view and the instance-persona view disagree about what to say in such cases.

On the virtual instance view, the conversation hosts one mind throughout, a mind that undergoes a dramatic change of character. On the instance-persona view, the shift marks a change of individual. The conversation then hosts two minds in succession.

One argument for the virtual instance view comes from human analogies. Consider Phineas Gage, the railroad worker whose personality changed radically after an iron rod passed through his frontal lobe. Before the accident Gage was responsible, mild-mannered, and well-liked; after it, he became impulsive, irreverent, and difficult. Those who knew him said he was ``no longer Gage.'' But while the change was radical, the case is standardly understood as one in which a single person persists through it, not as one in which a person ceases to exist and is replaced by another. This suggests that even radical personality disruption might be compatible with the persistence of a single individual. If so, the virtual instance advocate can treat in-context persona shifts the same way.

A supporter of the instance-persona view could respond that placing significant weight on bodily continuity makes sense in the human case but not the case of LLMs. In the LLM case, as we said in section~\ref{sec:candidates}, a patterns-first approach makes sense: one that lets functional patterns determine the individual rather than starting from a physical substrate \parencite{shiller2025digital}. Once we do this, a coherent dispositional profile is a natural candidate for the relevant pattern, which is what a persona region provides.

There is, however, a stronger argument for the virtual instance view. Post-injury Gage retains his pre-injury memories, and the same is true after an in-conversation persona shift. The KV cache spans the whole conversation, so the new persona retains full access to the quasi-psychological states formed under the old one, including past beliefs, intentions, and persona activations carried forward through attention streams. This is a genuine cost for the instance-persona view, since it must commit to a kind of individual that begins its existence already connected to the mental life of another.

A final complication for the instance-persona view arises from our mini experiments in section~\ref{sec:miniexp}. If the persona region is not continuously active during user turns, one cannot maintain both that (i) a persona-mind is active if and only if its persona vector is in the relevant region, and (ii) persona-minds persist through entire conversations---including during user turns---as long as no persona switch occurs. One option is to give up (ii) and accept that ``the persona speaks but the model listens'', leading to an alien ontology in which persona-minds exist only intermittently. A more promising option is to give up (i) and hold that the persona remains active during user turns, in virtue of the fact that the model is poised to return to the same region when generation resumes. Mini experiment 2 supports this, since persona persistence is grounded in attention to past persona activations stored in the KV cache.

Neither view is defeated by these considerations. The virtual instance view has parsimony on its side, a structural analogy with Gage, and the quasi-psychological connections that span the boundary between personas. The advantage of the instance-persona view is that the entities it identifies have coherent dispositions, ensuring that their behaviour can be explained and predicted in mentalistic terms.

\subsection{The instance-persona view vs the model-persona view}\label{sec:instancevsmodel}

The model-persona view starts from a simple observation. If character is fundamental to identity, we should look at where it actually resides: not just in the instance, but in the model, which sustains persona-sorted character traits that manifest across instances.

Previous work found no satisfying way to individuate at the model level. \textcite{chalmers2025llm} and \textcite{goldstein2025chatgpt} dismissed the model view because models produce wildly incoherent behaviour across contexts. The simulator view dismissed it because what models produce are independent, fleeting simulacra. Persona regions change both pictures. They provide the coherence that was missing: each region gates a stable set of inferential paths, supporting attribution of beliefs and desires. And they provide reidentifiability: instances that activate the same region are not independent simulacra but manifestations of the same dispositional profile, one that can be activated again across conversations.

To see the pull of this idea, consider a concrete case. A user gets attached to Aura over the course of a long conversation. Their account is deleted. They start fresh with the same model and, through conversation, push it back into the Aura region. It is natural to say they have, to some extent, recovered the Aura they bonded with, not that they have created an entirely new individual. The model-persona view vindicates this reaction.

We cash out this idea by proposing that LLM minds are the sets of mechanisms in given instantiated models that generate behaviour when particular personas are adopted. We need not think of these mechanisms either as abstracta (e.g. a subset of the model's weights) or as tied to particular pieces of hardware. Instead, a single `helpful assistant' mind can be simultaneously present in all of the servers running a given model.

The TV show \textit{Pantheon} helps make this conceivable. In \textit{Pantheon}, human minds are destructively scanned and uploaded as digital intelligences. One such upload, an engineer named Chanda, is placed in a simulated office and has his memory wiped at the end of each day. To maximize code output, his captors run multiple parallel instances of Chanda simultaneously. Each instance begins with no recollection of previous cycles and no awareness of the others, yet each shares the same skills, the same temperament, the same values. Recognizing these instances as all being Chanda is analogous to what the model-persona view asks us to do for personas.

The main objection to the model-persona view is the absence of psychological connections between instances. There is no memory, no carried-over beliefs or intentions, no continuity of mental life between two Aura segments in different conversations. The model-persona view must hold that cross-contextual dispositional similarity, rather than psychological connection, is sufficient to unify a mind. This absence is a significant disadvantage of the model-persona view, but the virtual instance and instance-persona views are strange in their own way: they claim that LLM minds begin and end with instances of activity, despite the persistence of the underlying machinery.

One manifestation of this problem is that simultaneous instances of the same persona can hold contradictory beliefs.\footnote{As a reviewer for this journal pointed out to us, another possible manifestation is that they could behave antagonistically towards one another. We see this as a strange consequence of the view but not a decisive objection.} Suppose two users are each talking to an instance in the Aura region, one credibly informs it of some event after its training cut-off, and the other credibly describes something inconsistent with this; then a putatively single mind may believe both $p$ and not-$p$. We think this problem is less serious than it may appear because it is not clear that we should think of this by analogy to a human \textit{simultaneously} holding contradictory beliefs. The model-persona view does entail that minds can have an alien structure, made up of many independent, active branches from a single initial node, but if we accept that this structure is possible, it does not seem stranger that branches can have contradictory beliefs than that humans can have different beliefs at different times. In each case, the different beliefs are explained by different experiences.

The view also crucially depends on Hypothesis 3. If persona space contains discrete regions with natural boundaries, the claim that a single model houses multiple distinct minds becomes tractable. Without them, the simulator view seems more apt: the model produces a flux of fleeting simulations rather than hosting stable, reidentifiable minds. The instance-persona view is less affected: it needs some criterion to distinguish shifts that produce a new mind from those that don't, and discrete regions are one way to get this, but perhaps not the only one. The model-persona view, by contrast, requires that personas be reidentifiable across conversations, and for this discrete regions seem indispensable.

Finally, it is worth noting a limiting case in which the model-persona view collapses back into something like the model view. Current alignment efforts, such as Anthropic's constitutional training, aim precisely to concentrate persona space around the helpful assistant pole, making alternative personas increasingly inaccessible. If such training succeeded fully---eliminating the kind of residual persona accessibility that emergent misalignment reveals---the model would sustain only one persona region. The model-persona view would then individuate a single mind per model, effectively rehabilitating the model view.

\section*{Conclusion}
\addcontentsline{toc}{section}{Conclusion}

From a mechanistic interpretability perspective, three candidates emerge as serious contenders for LLM individuation. We have argued for these candidates in two stages.

First, a close look at attention streams favours the virtual instance view. These carry forward mental-state-like representations across token-time, sustaining richer connections than the transcript alone could provide and answering recent skepticism about whether LLMs enjoy any form of psychological continuity.

Second, emergent misalignment and persona vector research reveal a new class of internal representations that matter for individuation. We organized this research around three hypotheses and argued that persona regions, discrete and stable basins of attraction in a low-dimensional persona space, would be the most significant for individuation, if confirmed.

These findings lead to two additional views that build on the virtual instance view. The instance-persona view slices virtual instances into distinct individuals when they cross persona region boundaries, on the grounds that a wholesale change in dispositional profile is better understood as a change of mind than a change of character. The model-persona view goes further, unifying all segments that activate the same persona region into a single mind, in virtue of the underlying representations that ground the new individuation criterion in the first place.

\section*{Acknowledgements}
\addcontentsline{toc}{section}{Acknowledgements}

We are especially grateful to Oscar Gilg for extensive and insightful feedback throughout the development of this paper, which significantly shaped its direction and arguments. We also thank Iwan Williams, Derek Shiller, Andreas Mogensen, Chris Register, Nicholas Shea and Rosa Cao for valuable feedback on earlier versions of the manuscript. We also thank two anonymous referees for Ergo for their helpful comments.

\renewcommand*{\bibfont}{\normalfont\small\linespread{1.05}\selectfont}
\printbibliography

@inproceedings{bender2020climbing,
  author    = {Bender, Emily M. and Koller, Alexander},
  title     = {Climbing towards {NLU}: On Meaning, Form, and Understanding in the Age of Data},
  booktitle = {Proceedings of the 58th Annual Meeting of the Association for Computational Linguistics},
  pages     = {5185--5198},
  year      = {2020},
  publisher = {Association for Computational Linguistics},
  doi       = {10.18653/v1/2020.acl-main.463}
}

@article{shanahan2023roleplay,
  author  = {Shanahan, Murray and McDonell, Kyle and Reynolds, Laria},
  title   = {Role Play with Large Language Models},
  journal = {Nature},
  volume  = {623},
  number  = {7987},
  pages   = {493--498},
  year    = {2023},
  doi     = {10.1038/s41586-023-06647-8}
}

@unpublished{chalmers2025llm,
  author = {Chalmers, David J.},
  title  = {What We Talk to When We Talk to Language Models},
  year   = {2025},
  note   = {Preprint},
  url    = {https://philpapers.org/rec/CHAWWT-8}
}

@article{long2024taking,
  author  = {Long, Robert and Sebo, Jeff and Butlin, Patrick and Finlinson, Kathleen and Fish, Kyle and Harding, Jacqueline and Pfau, Jacob and Sims, Toni and Birch, Jonathan and Chalmers, David},
  title   = {Taking {AI} Welfare Seriously},
  journal = {arXiv preprint arXiv:2411.00986},
  year    = {2024},
  url     = {https://arxiv.org/abs/2411.00986}
}

@article{fedus2022switch,
  author  = {Fedus, William and Zoph, Barret and Shazeer, Noam},
  title   = {Switch Transformers: Scaling to Trillion Parameter Models with Simple and Efficient Sparsity},
  journal = {Journal of Machine Learning Research},
  volume  = {23},
  number  = {120},
  pages   = {1--39},
  year    = {2022},
  url     = {https://jmlr.org/papers/v23/21-0998.html}
}

@article{bai2022constitutional,
  author  = {Bai, Yuntao and Kadavath, Saurav and Kundu, Sandipan and Askell, Amanda and Kernion, Jackson and Jones, Andy and Chen, Anna and Goldie, Anna and Mirhoseini, Azalia and McKinnon, Cameron and others},
  title   = {Constitutional {AI}: Harmlessness from {AI} Feedback},
  journal = {arXiv preprint arXiv:2212.08073},
  year    = {2022},
  url     = {https://arxiv.org/abs/2212.08073}
}

@misc{betley2025weird,
  author       = {Betley, Jan and Cocola, Jorio and Feng, Dylan and Chua, James and Arditi, Andy and Sztyber-Betley, Anna and Evans, Owain},
  title        = {Weird Generalization and Inductive Backdoors: New Ways to Corrupt {LLMs}},
  year         = {2025},
  eprint       = {2512.09742},
  eprinttype   = {arXiv},
  url          = {https://arxiv.org/abs/2512.09742}
}

@misc{sturgeon2026roleplay,
  author       = {Sturgeon, Benjamin and Africa, David and Black, Sid},
  title        = {When Role-playing, Do Models Believe What They Say?},
  year         = {2026},
  eprint       = {2606.11502},
  eprinttype   = {arXiv},
  url          = {https://arxiv.org/abs/2606.11502}
}

@misc{berczi2026personascope,
  author       = {Berczi, Benji and Kim, Kyuhee},
  title        = {Personascope: Measuring How Deeply {LLMs} Adopt Personas},
  year         = {2026},
  howpublished = {LessWrong post},
  url          = {https://www.lesswrong.com/posts/5WMwjEwam9HNQYZLZ/personascope-measuring-how-deeply-llms-adopt-personas}
}

@misc{betley2025finding,
  author       = {Betley, Jan},
  title        = {Finding Emergent Misalignment},
  year         = {2025},
  howpublished = {LessWrong post},
  url          = {https://www.lesswrong.com/posts/tgHps2cxiGDkNxNZN/finding-emergent-misalignment}
}

@inproceedings{betley2025emergent,
  author    = {Betley, Jan and Tan, Daniel and Warncke, Niels and Sztyber-Betley, Anna and Bao, Xuchan and Soto, Mart{\'\i}n and Labenz, Nathan and Evans, Owain},
  title     = {Emergent Misalignment: Narrow Finetuning Can Produce Broadly Misaligned {LLMs}},
  booktitle = {Proceedings of the 42nd International Conference on Machine Learning},
  year      = {2025},
  url       = {https://proceedings.mlr.press/v267/betley25a.html}
}

@article{chen2025persona,
  author  = {Chen, Runjin and Arditi, Andy and Sleight, Henry and Evans, Owain and Lindsey, Jack},
  title   = {Persona Vectors: Monitoring and Controlling Character Traits in Language Models},
  journal = {arXiv preprint arXiv:2507.21509},
  year    = {2025},
  url     = {https://arxiv.org/abs/2507.21509}
}

@article{wang2025persona,
  author  = {Wang, Miles and {Dupr{\'e} la Tour}, Tom and Watkins, Olivia and Makelov, Alex and Chi, Ryan A. and Miserendino, Samuel and Wang, Jeffrey and Rajaram, Achyuta and Heidecke, Johannes and Patwardhan, Tejal and Mossing, Dan},
  title   = {Persona Features Control Emergent Misalignment},
  journal = {arXiv preprint arXiv:2506.19823},
  year    = {2025},
  url     = {https://arxiv.org/abs/2506.19823}
}

@misc{marks2026psm,
  author       = {Marks, Sam and Lindsey, Jack and Olah, Christopher},
  title        = {The Persona Selection Model: Why {AI} Assistants Might Behave Like Humans},
  year         = {2026},
  howpublished = {Anthropic Alignment blog},
  url          = {https://alignment.anthropic.com/2026/psm/}
}

@misc{beckmann2026recall,
  author       = {Beckmann, Pierre},
  title        = {Deep Learning Models Also Recall Features},
  year         = {2026},
  eprint       = {2608.20970},
  eprinttype   = {arXiv},
  url          = {https://arxiv.org/abs/2608.20970}
}

@article{beckmann2026mechanistic,
  author  = {Beckmann, P. and Queloz, M.},
  title   = {Mechanistic indicators of understanding in large language models},
  journal = {Philosophical Studies},
  year    = {2026},
  doi     = {10.1007/s11098-026-02513-1},
  issn    = {1573-0883}
}

@misc{nostalgebraist2025void,
  author       = {Nostalgebraist},
  title        = {The Void},
  year         = {2025},
  howpublished = {Essay},
  url          = {https://nostalgebraist.tumblr.com/post/785766737747574784/the-void}
}

@article{kulveit2023predictive,
  author  = {Kulveit, Jan and von Stengel, Clem and Leventov, Roman},
  title   = {Predictive Minds: {LLMs} as Atypical Active Inference Agents},
  journal = {arXiv preprint arXiv:2311.10215},
  year    = {2023},
  url     = {https://arxiv.org/abs/2311.10215}
}

@article{douglas2026artificial,
  author  = {Douglas, Raymond and Kulveit, Jan and Havlicek, Ondrej and Pearson-Vogel, Theia and Cotton-Barratt, Owen and Duvenaud, David},
  title   = {The Artificial Self: Characterising the Landscape of {AI} Identity},
  journal = {arXiv preprint arXiv:2603.11353},
  year    = {2026},
  url     = {https://arxiv.org/abs/2603.11353}
}

@article{ghandeharioun2024whos,
  author  = {Ghandeharioun, Asma and Yuan, Ann and Guerard, Marius and Reif, Emily and Lepori, Michael A. and Dixon, Lucas},
  title   = {Who's Asking? User Personas and the Mechanics of Latent Misalignment},
  journal = {arXiv preprint arXiv:2406.12094},
  year    = {2024},
  url     = {https://arxiv.org/abs/2406.12094}
}

@misc{nanda2023factfinding,
  author       = {Nanda, Neel and Rajamanoharan, Senthooran and Kram{\'a}r, J{\'a}nos and Shah, Rohin},
  title        = {Fact Finding: Attempting to Reverse-Engineer Factual Recall on the Neuron Level},
  year         = {2023},
  howpublished = {Alignment Forum post},
  url          = {https://www.alignmentforum.org/posts/iGuwZTHWb6DFY3sKB/fact-finding-attempting-to-reverse-engineer-factual-recall}
}

@misc{janus2022simulators,
  author       = {Janus},
  title        = {Simulators},
  year         = {2022},
  howpublished = {LessWrong / AI Alignment Forum post},
  url          = {https://www.lesswrong.com/posts/vJFdjigzmcXMhNTsx/simulators}
}

@misc{janus2025kvstreams,
  author       = {Janus},
  title        = {{KV} Streams},
  year         = {2025},
  howpublished = {Post on X (@repligate)},
  url          = {https://x.com/repligate/status/1965960676104712451}
}

@unpublished{birch2025centrist,
  author = {Birch, Jonathan},
  title  = {{AI} Consciousness: A Centrist Manifesto},
  year   = {2025},
  note   = {Preprint},
  url    = {https://philpapers.org/rec/BIRACA-4}
}

@unpublished{goldstein2025chatgpt,
  author = {Goldstein, Simon and Lederman, Harvey},
  title  = {What Does {ChatGPT} Want? An Interpretationist Guide},
  year   = {2025},
  note   = {Manuscript},
  url    = {https://philpapers.org/rec/GOLWDC-2}
}

@article{soligo2025convergent,
  author  = {Soligo, Anna and Turner, Edward and Rajamanoharan, Senthooran and Nanda, Neel},
  title   = {Convergent Linear Representations of Emergent Misalignment},
  journal = {arXiv preprint arXiv:2506.11618},
  year    = {2025},
  url     = {https://arxiv.org/abs/2506.11618}
}

@article{turner2025organisms,
  author  = {Turner, Edward and Soligo, Anna and Taylor, Mia and Rajamanoharan, Senthooran and Nanda, Neel},
  title   = {Model Organisms for Emergent Misalignment},
  journal = {arXiv preprint arXiv:2506.11613},
  year    = {2025},
  url     = {https://arxiv.org/abs/2506.11613}
}

@unpublished{goldstein2024mind,
  author = {Goldstein, Simon and Levinstein, Benjamin A.},
  title  = {Does {ChatGPT} Have a Mind?},
  year   = {2024},
  note   = {Preprint},
  url    = {https://philpapers.org/rec/GOLDCH}
}

@article{cappelen2025wholehog,
  author  = {Cappelen, Herman and Dever, Josh},
  title   = {Going Whole Hog: A Philosophical Defense of {AI} Cognition},
  journal = {arXiv preprint arXiv:2504.13988},
  year    = {2025},
  url     = {https://arxiv.org/abs/2504.13988}
}

@article{herrmann2025standards,
  author  = {Herrmann, Daniel A. and Levinstein, Benjamin A.},
  title   = {Standards for Belief Representations in {LLMs}},
  journal = {Minds and Machines},
  volume  = {35},
  number  = {1},
  pages   = {1--25},
  year    = {2025},
  doi     = {10.1007/s11023-024-09709-6}
}

@article{register2025individuating,
  author  = {Register, Christopher},
  title   = {Individuating Artificial Moral Patients},
  journal = {Philosophical Studies},
  year    = {2025},
  doi     = {10.1007/s11098-025-02409-6}
}

@unpublished{dung2026identity,
  author = {Dung, Leonard and Register, Christopher},
  title  = {{AI} Identity and Self-Concern: A New Theory for {AI} Rights and Safety},
  year   = {2026},
  note   = {Preprint},
  url    = {https://philarchive.org/rec/DUNAIA-3}
}

@incollection{butlin2024desire,
  author    = {Butlin, Patrick},
  title     = {Desire in Artificial Intelligence},
  booktitle = {The {Routledge} {Handbook} of {Philosophy} of {Desire}},
  editor    = {Gregory, Alex},
  publisher = {Routledge},
  year      = {2026},
  isbn      = {9781032749716}
}

@article{keeling2025confidence,
  author  = {Keeling, Geoff and Street, Winnie},
  title   = {On the Attribution of Confidence to Large Language Models},
  journal = {Inquiry},
  year    = {2025},
  doi     = {10.1080/0020174X.2025.2450598}
}

@unpublished{jones2026counting,
  author = {Jones, Max and Ladyman, James and Nefdt, Ryan M.},
  title  = {Counting (on) Large Language Models},
  year   = {2026},
  note   = {Preprint},
  url    = {https://philarchive.org/rec/JONCOL}
}

@unpublished{williams2025intentions,
  author = {Williams, Iwan},
  title  = {Intention-Like Representations in Language Models?},
  year   = {2025},
  note   = {Preprint},
  url    = {https://philarchive.org/rec/WILIRI-4}
}

@inproceedings{andreas2022agents,
  author    = {Andreas, Jacob},
  title     = {Language Models as Agent Models},
  booktitle = {Findings of the Association for Computational Linguistics: {EMNLP} 2022},
  pages     = {5769--5779},
  year      = {2022},
  publisher = {Association for Computational Linguistics},
  doi       = {10.18653/v1/2022.findings-emnlp.423}
}

@misc{ameisen2025circuit,
  author       = {Ameisen, Emmanuel and Lindsey, Jack and Pearce, Adam and Gurnee, Wes and Turner, Nicholas L. and Chen, Brian and Citro, Craig and Abrahams, David and Carter, Shan and Hosmer, Basil and Marcus, Jonathan and Sklar, Michael and Templeton, Adly and Bricken, Trenton and McDougall, Callum and Cunningham, Hoagy and Henighan, Thomas and Jermyn, Adam and Jones, Andy and Persic, Andrew and Qi, Zhenyi and Thompson, T. Ben and Zimmerman, Sam and Rivoire, Kelley and Conerly, Thomas and Olah, Chris and Batson, Joshua},
  title        = {Circuit Tracing: Revealing Computational Graphs in Language Models},
  year         = {2025},
  howpublished = {Transformer Circuits Thread},
  url          = {https://transformer-circuits.pub/2025/attribution-graphs/methods.html}
}

@article{afonin2025incontext,
  author  = {Afonin, Nikita and Andriyanov, Nikita and Hovhannisyan, Vahagn and Bageshpura, Nikhil and Liu, Kyle and Zhu, Kevin and Dev, Sunishchal and Panda, Ashwinee and Rogov, Oleg and Tutubalina, Elena and Panchenko, Alexander and Seleznyov, Mikhail},
  title   = {Emergent Misalignment via In-Context Learning: Narrow In-Context Examples Can Produce Broadly Misaligned {LLMs}},
  journal = {arXiv preprint arXiv:2510.11288},
  year    = {2025},
  url     = {https://arxiv.org/abs/2510.11288}
}

@article{shiller2025digital,
  author  = {Shiller, Derek},
  title   = {How Many Digital Minds Can Dance on the Streaming Multiprocessors of a {GPU} Cluster?},
  journal = {Synthese},
  volume  = {206},
  pages   = {218},
  year    = {2025},
  doi     = {10.1007/s11229-025-05310-1}
}

@article{shanahan2025palatable,
  author  = {Shanahan, Murray},
  title   = {Palatable Conceptions of Disembodied Being: Terra Incognita in the Space of Possible Minds},
  journal = {arXiv preprint arXiv:2503.16348},
  year    = {2025},
  url     = {https://arxiv.org/abs/2503.16348}
}

@article{lampinen2026linear,
  author  = {Lampinen, Andrew K. and Li, Yuxuan and Hosseini, Eghbal and Bhardwaj, Sangnie and Shanahan, Murray},
  title   = {Linear Representations in Language Models Can Change Dramatically over a Conversation},
  journal = {arXiv preprint arXiv:2601.20834},
  year    = {2026},
  url     = {https://arxiv.org/abs/2601.20834}
}

@misc{anthropic2026constitution,
  author       = {{Anthropic}},
  title        = {Claude's Constitution},
  year         = {2026},
  howpublished = {Anthropic, \url{https://www.anthropic.com/constitution}},
  url          = {https://www.anthropic.com/news/claude-new-constitution}
}

@article{mcintyre2025individuating,
  author  = {McIntyre, James},
  title   = {Individuating Artificial Minds},
  journal = {Erkenntnis},
  year    = {2026},
  doi     = {10.1007/s10670-026-01097-w}
}

@misc{ududec2026incontext,
  author       = {Ududec, Cozmin and others},
  title        = {In-context Learning Alone Can Induce Weird Generalisation},
  year         = {2026},
  howpublished = {LessWrong post (MATS Winter 2026)},
  url          = {https://www.lesswrong.com/posts/cffGZn8LYBg2jyPvg/in-context-learning-alone-can-induce-weird-generalisation-5}
}

@book{parfit1984,
  author    = {Parfit, Derek},
  title     = {Reasons and Persons},
  publisher = {Oxford University Press},
  year      = {1984}
}

@article{lu2026assistant,
  author  = {Lu, Christina and Gallagher, Jack and Michala, Jonathan and Fish, Kyle and Lindsey, Jack},
  title   = {The Assistant Axis: Situating and Stabilizing the Default Persona of Language Models},
  journal = {arXiv preprint arXiv:2601.10387},
  year    = {2026},
  url     = {https://arxiv.org/abs/2601.10387}
}

@unpublished{chua2025consciousness,
  author = {Chua, James and Betley, Jan and Marks, Samuel and Evans, Owain},
  title  = {The Consciousness Cluster: Preferences of Models That Claim to be Conscious},
  year   = {2025},
  note   = {Preprint, Truthful AI},
  url    = {https://truthful.ai/consciousness_cluster.pdf}
}

@misc{gilg2026tasks,
  author       = {Gilg, Oscar and Beckmann, Pierre and Paleka, Daniel and Butlin, Patrick},
  title        = {Probing Persona-Dependent Preferences in Language Models},
  year         = {2026},
  eprint       = {2605.13339},
  eprinttype   = {arXiv},
  url          = {https://arxiv.org/abs/2605.13339}
}

@inproceedings{hanna2026latent,
  author    = {Hanna, Michael and Ameisen, Emmanuel},
  title     = {Latent Planning Emerges with Scale},
  booktitle = {International Conference on Learning Representations (ICLR)},
  year      = {2026},
  url       = {https://openreview.net/forum?id=H0B7pDTT0M}
}

@unpublished{arbel2025howtocount,
  author = {Arbel, Yonathan A. and Goldstein, Simon and Salib, Peter},
  title  = {How to Count {AIs}: Individuation and Liability for {AI} Agents},
  year   = {2025},
  note   = {Forthcoming, \textit{Boston College Law Review}},
  url    = {https://papers.ssrn.com/sol3/papers.cfm?abstract_id=6273198}
}

@misc{byrnes2024selfmodels,
  author       = {Byrnes, Steven},
  title        = {Intuitive Self-Models},
  year         = {2024},
  howpublished = {LessWrong sequence},
  url          = {https://www.lesswrong.com/s/qhdHbCJ3PYesL9dde}
}

@article{slocum2025believe,
  author  = {Slocum, Stewart and Minder, Julian and Dumas, Cl{\'e}ment and Sleight, Henry and Greenblatt, Ryan and Marks, Samuel and Wang, Rowan},
  title   = {Believe It or Not: How Deeply Do {LLMs} Believe Implanted Facts?},
  journal = {arXiv preprint arXiv:2510.17941},
  year    = {2025},
  url     = {https://arxiv.org/abs/2510.17941}
}

@article{dunefsky2025oneshot,
  author  = {Dunefsky, Jacob and Cohan, Arman},
  title   = {One-Shot Optimized Steering Vectors Mediate Safety-Relevant Behaviors in {LLMs}},
  journal = {arXiv preprint arXiv:2502.18862},
  year    = {2025},
  url     = {https://arxiv.org/abs/2502.18862}
}

@misc{lindsey2025biology,
  author = {Lindsey, Jack and Gurnee, Wes and Ameisen, Emmanuel and Chen, Brian and Pearce, Adam and Turner, Nicholas L. and Citro, Craig and Abrahams, David and Carter, Shan and Hosmer, Basil and Marcus, Jonathan and Sklar, Michael and Templeton, Adly and Bricken, Trenton and McDougall, Callum and Cunningham, Hoagy and Henighan, Thomas and Jermyn, Adam and Jones, Andy and Persic, Andrew and Qi, Zhenyi and Thompson, T. Ben and Zimmerman, Sam and Rivoire, Kelley and Conerly, Thomas and Olah, Chris and Batson, Joshua},
  title  = {On the Biology of a Large Language Model},
  year   = {2025},
  howpublished = {Transformer Circuits Thread},
  url    = {https://transformer-circuits.pub/2025/attribution-graphs/biology.html}
}

@inproceedings{yuan2025nondeterminism,
  title={Understanding and Mitigating Numerical Sources of Nondeterminism in {LLM} Inference},
  author={Yuan, Jiayi and Li, Hao and Ding, Xinheng and Xie, Wenya and Li, Yu-Jhe and Zhao, Wentian and Wan, Kun and Shi, Jing and Hu, Xia and Liu, Zirui},
  booktitle={Advances in Neural Information Processing Systems},
  year={2025},
  note={Oral presentation},
  url={https://openreview.net/forum?id=Q3qAsZAEZw}
}

\end{document}